%% file: main.tex
\pdfoutput=1
\documentclass[11pt]{article}

\usepackage{arxiv}

\usepackage[utf8]{inputenc}
\usepackage[T1]{fontenc}
\usepackage{hyperref}
\usepackage{url}
\usepackage{booktabs}
\usepackage{amsfonts}
\usepackage{nicefrac}
\usepackage{microtype}
\usepackage{xcolor}
\usepackage{amsmath}
\usepackage{graphicx}
\usepackage{algorithm}
\usepackage{algorithmic}
\usepackage{xspace}
\usepackage{amssymb}      
\usepackage{natbib}         
\usepackage{listings}
\usepackage{enumitem}    
\usepackage{tcolorbox}
\usepackage{tabularx}
\usepackage{multirow}
\usepackage{float}
\usepackage{pgfplots}
\usepackage{CJKutf8}        
\usepackage{tikz}
\usepackage[table]{xcolor}
\usepackage{caption}
\usepackage{lineno}
\usetikzlibrary{arrows.meta, positioning, matrix, decorations.pathreplacing, calc}
\pgfplotsset{compat=1.18}

\tcbuselibrary{breakable}
\tcbuselibrary{skins}

\tcbset{
  casebox/.style={
    enhanced,
    colframe=gray!50, 
    arc=3pt, 
    boxrule=0.8pt,
    left=8pt, right=8pt, top=10pt, bottom=10pt,
    fonttitle=\bfseries\small, 
    coltitle=black,
    attach boxed title to top left={xshift=4mm, yshift=-3mm},
    boxed title style={colback=white, frame hidden},
    colback=#1 
  }
}

\lstdefinelanguage{json}{
  morestring=[b]",%
  morecomment=[l]{//},
  morekeywords={true,false,null},
  sensitive=false,
}

\newcommand{\framework}{InfoSeeker\xspace}

\title{\framework: A Scalable Hierarchical Parallel Agent Framework for Web Information Seeking}

\usepackage{authblk}            
\usepackage{hyperref}           
\usepackage[symbol]{footmisc}   
\newcommand{\equalmark}{\textsuperscript{†}}   
\makeatletter
\renewcommand\AB@affilnote[1]{}  
\makeatother
\author[1]{Ka Yiu~Lee\equalmark}
\author[2]{Yuxuan~Huang\equalmark}
\author[1]{Zhiyuan~He\equalmark}
\author[3]{Huichi~Zhou\equalmark}
\author[1]{Weilin~Luo}
\author[1]{Kun~Shao}
\author[2]{Meng~Fang}
\author[3]{Jun~Wang}
\affil[]{\parbox{\textwidth}{
  \centering \textsuperscript{1}\,Huawei Noah's Ark Lab \quad \textsuperscript{2}\,University of Liverpool   \quad  \textsuperscript{3}\,University College London}}

\begin{document}

\maketitle

\begin{abstract}
Recent agentic search systems have made substantial progress by emphasising deep, multi-step reasoning. However, this focus often overlooks the challenges of wide-scale information synthesis, where agents must aggregate large volumes of heterogeneous evidence across many sources. As a result, most existing large language model agent systems face severe limitations in data-intensive settings, including context saturation, cascading error propagation, and high end-to-end latency.
To address these challenges, we present \framework, a hierarchical framework based on principle of near-decomposability, containing a strategic \textit{Host}, multiple \textit{Managers} and parallel \textit{Workers}. By leveraging aggregation and reflection mechanisms at the Manager layer, our framework enforces strict context isolation to prevent saturation and error propagation. Simultaneously, the parallelism in worker layer accelerates the speed of overall task execution, mitigating the significant latency. Our evaluation on two complementary benchmarks demonstrates both efficiency ($ 3-5 \times$ speed-up) and effectiveness, achieving a $8.4\%$ success rate on WideSearch-en and $52.9\%$ accuracy on BrowseComp-zh.  The
code is released at \url{https://github.com/agent-on-the-fly/InfoSeeker}.
\end{abstract}

\begin{figure}[H]
  \centering
  \includegraphics[width=0.75\linewidth]{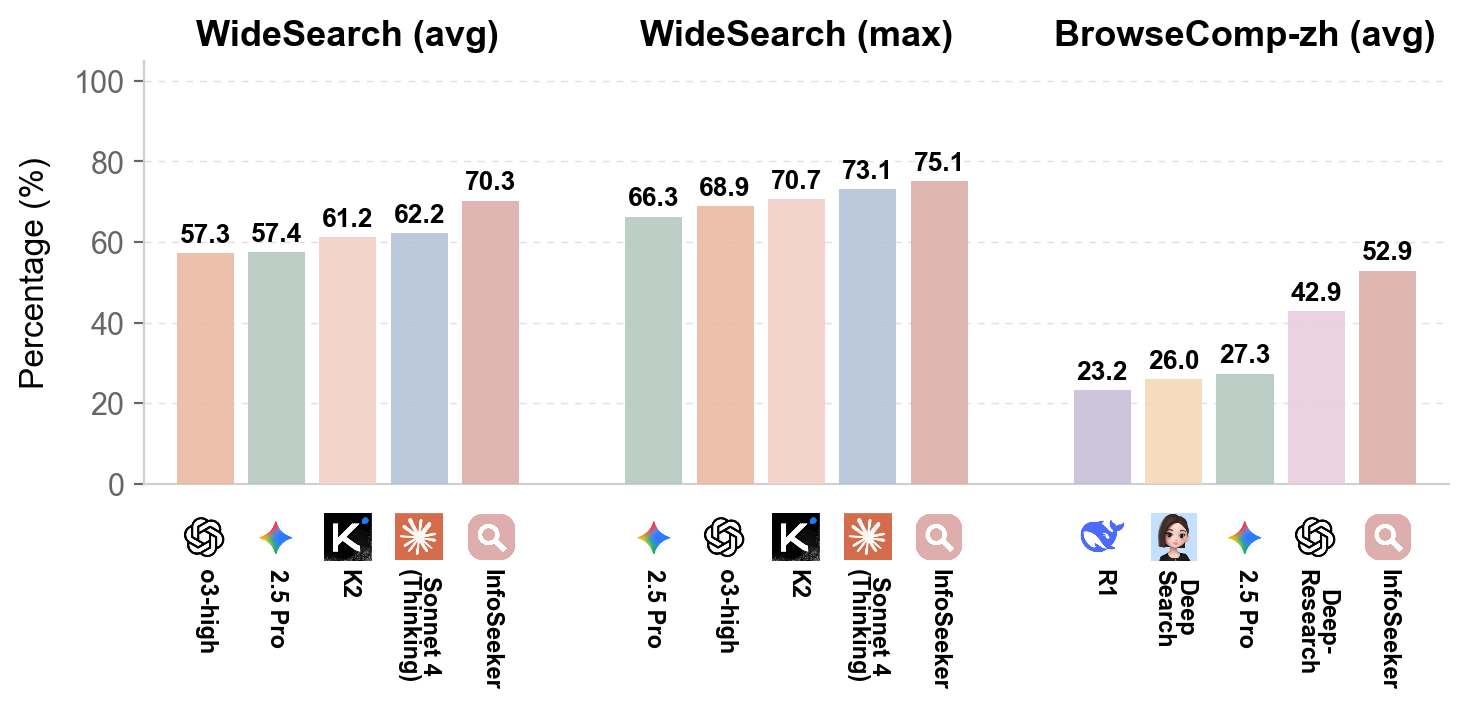}
  \caption{Performance results on \textit{BrowseComp-zh} (avg) and \textit{WideSearch} (avg/max).}
  \label{fig:bar}
\end{figure}

\input{sec/intro_v2}

\input{sec/rw}

\input{sec/method}

\input{sec/experiments}

\input{sec/discussion}

\input{sec/conclusion}

\input{sec/limitations}

\bibliographystyle{plainnat} 
\bibliography{main}

\appendix
\input{sec/appendix}

\end{document}

%% file: sec/intro_v2.tex
\section{Introduction}
\label{sec:intro}

As Large Language Models (LLMs) continue to evolve~\citep{zhao2023survey}, the paradigm of web search is shifting from simple information retrieval to autonomous agentic web search~\citep{yang2025agentic}. Users are no longer satisfied with simple multi-hop question answering. Instead, they require models capable of processing data-intensive and long-horizon tasks, such as Deep Research~\citep{huang2025deep}. Consequently, the community has largely focused on optimising agents for multi-step reasoning and complex logical chains~\citep{mialon2023gaia,wei2025browsecomp,zhou2025browsecomp}, witnessing a proliferation of architectures designed to maximise performance on these depth-oriented benchmarks~\citep{google2025deepresearch,openai2025deepresearch,hu2025owl}. 


However, benchmarks such as WideSearch reveals critical deficiencies in current LLM agents when applied to large-scale information seeking, specifically regarding incomplete planning, lack of reflection, and the misuse of retrieved evidence~\citep{wong2025widesearch}. We observed similar patterns in our pilot experiments: when a task requires aggregating data across dozens of web pages, the context windows of traditional frameworks (e.g., Gemini DeepResearch~\citep{google2025deepresearch}) saturate almost immediately, leading to failure. Meanwhile, sequential agent frameworks like MiroThinker~\citep{team2025mirothinker} and WebSailor~\citep{li2025websailor}, which rely on ReAct-style loops~\citep{yao2022react}, encounter severe error accumulation, as early mistakes tend to compound over time. Lastly, these sequential frameworks result in unacceptable time delays, severely limiting their efficiency.

Building on these observations, we argue that simply expanding context windows or model capacity cannot fundamentally resolve the challenges of wide-scale information seeking. Inspired by Near-decomposability~\citep{Simon1991}, which posits that complex systems function best when divided into semi-autonomous modules. In this structure, subsystems operate independently on details (short-run independence) and coordinate only through high-level summaries (long-run dependence). We propose \framework, a hierarchical framework comprising three distinct layers: a strategic \textit{Host} that maintains compressed global state and plans high-level directives, domain-specific \textit{Managers} that decompose these directives, verify quality, and aggregate results and a \textit{Worker} layer. Crucially, this design unlocks massively parallel execution at the \textit{Worker} level: multiple workers execute atomic tool interactions via the Model Context Protocol (MCP)~\citep{modelcontextprotocol}  simultaneously. By isolating these concurrent execution streams and propagating only concise summaries upward, our hierarchy effectively mitigates context saturation, error propagation and latency.

We empirically evaluate \framework on two complementary benchmarks: BrowseComp-zh (Depth \& Chinese)~\citep{zhou2025browsecomp} and WideSearch (Width \& English \& Chinese)~\citep{wong2025widesearch}. Results demonstrate that \framework achieves superior performance, securing a $66.7\%$ improvement in task success on WideSearch and a $13.8\%$ accuracy gain on BrowseComp-zh. Notably, it outperforms state-of-the-art commercial baselines, including Gemini Deep Research~\citep{google2025deepresearch} and OpenAI Deep Research~\citep{openai2025deepresearch}. Crucially, our parallel architecture also delivers significant efficiency gains, achieving an approximately $3-5 \times$ speed-up in inference latency compared to sequential execution.

%% file: sec/rw.tex
\section{Related Work}
\label{sec:related_work}

In recent years, agentic search~\citep{huang2025deep} has evolved alongside LLMs, moving from conventional web search and LLM-enhanced retrieval-augmented generation~\citep{press2022selfask,khattab2022dsp,gao2023rag} to search agents characterised by autonomous planning and multi-round querying~\citep{chen2024mindsearch,zhang2025webpilot}. Rather than limiting themselves to single-turn query rewriting, contemporary systems construct parallel, sequential, or hybrid search structures conditioned on user intent and environmental context~\citep{ahluwalia2024hybrid}. By iteratively refining queries in response to dynamic feedback~\citep{chan2024rq,madaan2023self}, these agents effectively achieve inference-time compute scaling through the test-time expansion of search.


\textbf{Agentic Workflow Orchestration.} Building upon foundational tool-augmented frameworks such as WebGPT \citep{nakano2021webgpt} and ReAct \citep{yao2022react}, contemporary deep research agents, notably GPT-Researcher \citep{gpt_researcher} and Open Deep Search \citep{open_deep_research}, decompose complex queries into granular subtasks, evaluated by emerging benchmarks like DeepResearchGym \citep{coelho2025deepresearchgym}. Orchestration has further evolved towards compiling goals into executable graphs via MCTS-guided search \citep{zhang2024aflow} or evolutionary workflows \citep{niu2025flow}. Whilst production-grade frameworks like AutoGen \citep{wu2024autogen} and LangGraph \citep{langgraph} provide essential abstractions, they primarily optimise workflows offline, offering limited runtime control over active graphs. Consequently, they lack support for real-time replanning and cross-branch compute reallocation.

\textbf{Parallel Reasoning and Execution.} Parallel execution is a vital lever for mitigating the latency of agentic search. At the token and action levels, speculative decoding~\citep{leviathan2023fast, miao2023specinfer, cai2024medusa} and speculative reasoning~\citep{pan2025specreason, yang2025speculative} reduce inference time via draft-and-verify and multi-token prediction. At the reasoning level, Dynamic Parallel Tree Search~\citep{ding2025dynamic} accelerates tree-of-thought exploration, whilst ParaThinker~\citep{wen2025parathinker} and Parallel-R1~\citep{zheng2025parallel} instil native parallel reasoning capabilities. Flash-Searcher~\citep{qin2025flash} and FlashResearch~\citep{nie2025flashresearch} further employ DAG-based execution and dynamic tree decomposition to parallelise complex sub-tasks. Whilst these approaches improve efficiency, they still assume static branching and fixed reasoning structures. \framework instead lifts parallelism to the workflow level with a three-tier hierarchy, enabling speculative branch expansion with real-time pruning or escalation. With a MapReduce-style pattern and MCP-based context isolation, it separates reasoning depth from execution width, scaling to long horizons within bounded contexts.

%% file: sec/method.tex
\section{Framework}
\label{sec:methodology}

\begin{figure*}[t]
  \centering
  \includegraphics[width=0.75\linewidth]{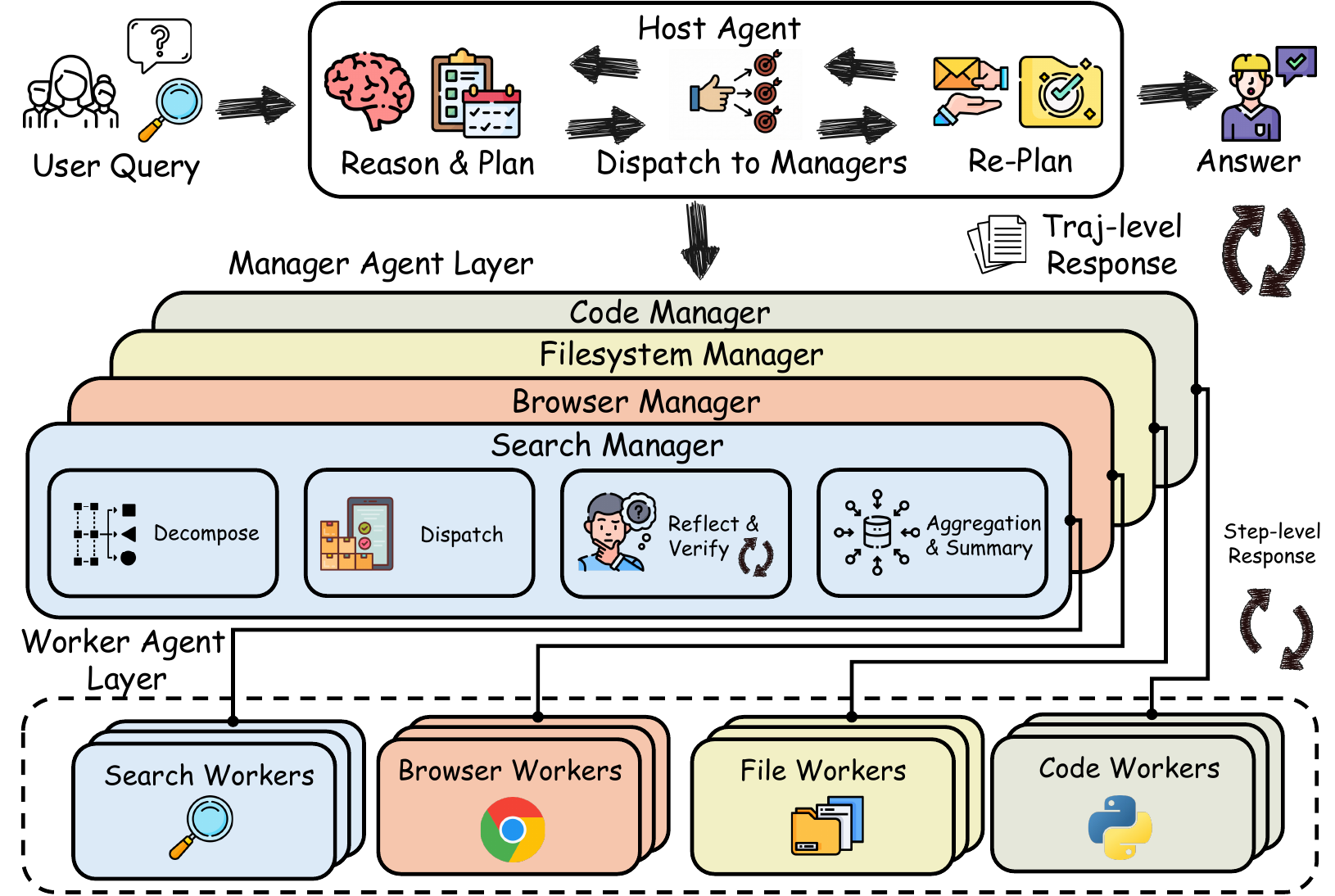}  
  \caption{\textbf{Overview of the \framework framework.} The system features a three-tier topology consisting of a strategic Host, domain-specific Managers, and tool-executing Workers. By enforcing hierarchical context isolation, high-level directives ($q_t$) are decomposed into parallelisable subtasks ($q_t^k$) by Managers and executed by Workers. Final results are aggregated into concise summaries ($y_t$) to support long-horizon planning while preventing context exhaustion at the strategic level.}
  \label{fig:framework-architecture}
\end{figure*}

We propose \framework, a hierarchical system based on near-decomposability. Figure~\ref{fig:framework-architecture} depicts its three-tier topology: a strategic \textit{Host}, \textit{Managers} for parallel orchestration, and \textit{Workers} for tool interaction. This design utilises the MCP to enforce context isolation, allowing the system to scale reasoning depth and execution width independently while maintaining bounded contexts.

\input{sec/code}

\subsection{Hierarchical Architecture}
\label{subsec:hierarchy}

\framework adopts a three-layer hierarchical architecture consisting of a host agent \(\hat{A} \), a set of domain-specific manager agents \( \{\tilde{A} \} \), and their associated worker agents \( \{\bar{A} \} \). Algorithm~\ref{alg:pseudocode} summarises the end-to-end execution workflow of \framework, illustrating how an initial user query is sequentially decomposed by the host, executed through domain-specific managers and their worker pools, and progressively abstracted into concise results that guide subsequent planning steps.

\paragraph{Host agent.}
Given an initial user query \( Q \), the host agent \(\hat{A} \) performs high-level, sequential reasoning. It maintains a context

\begin{equation}
\hat{C}_{t-1} = (Q, q_0, y_0, q_1, y_1, \dots, q_{t-1}, y_{t-1}),
\end{equation}

where each \( q_i \) denotes a high-level step generated by the host and each \( y_i \) is the corresponding response returned by a manager. At iteration \(t\), conditioned on the current context, the host generates the next step and if execution is not finished, a manager \(\tilde{A}_{t}\) is selected to solve \( q_t \) and returns a step-level result \( y_t \),

\begin{align}
(q_t, \tilde{A}_{t}) 
&\leftarrow \hat{A}\!\left(\hat{C}_{t-1}\right), \\
y_t 
&\leftarrow \tilde{A}_{t}(q_t).
\end{align}
 after which the host updates its context as

\begin{equation}
\hat{C}_{t} 
\leftarrow \hat{C}_{t-1} \oplus (q_t, y_t).
\end{equation}
The host never accesses subtasks, tool calls, or intermediate execution details. Its reasoning is strictly based on the sequence of step–response pairs, which bounds host-level context growth and enables long-horizon planning.

When the host agent determines that execution should terminate at iteration \(T\), it produces the final output by leveraging the complete accumulated context \(\hat{C}_T\). Formally, the final response is produced as
\begin{equation}
y^\ast \leftarrow \hat{A}(\hat{C}_T),
\end{equation}
where \(y^\ast\) denotes the final answer returned to the user. This design allows the host to make a global, informed decision based on all preceding step while response pairs remaining isolated from managers’ internal reasoning and tool-level executions.

\paragraph{Manager agents.}
\(\tilde{A}_{t}\) is specialised to a particular domain such as web search. Upon receiving a step \( q_t \), the manager dynamically decomposes it into a set of \( P_t \) parallelisable subtasks
\begin{equation}
    \{q^{1}_{t}, q^{2}_{t}, \dots, q^{P_t}_{t}\}= \text{Decomp}_{\tilde{A}_{t}}(q_t),
\end{equation}

which are dispatched concurrently to worker agents. The manager oversees each execution, performs validation or revision if necessary, and aggregates the final worker results into a step-level response:

\begin{equation}
\begin{aligned}
y_t
&= \tilde{A} _{t}(q_t) \\
&= \text{Aggr}_{\tilde{A} _{t}}\!\Big(
\{\bar{A}^{k}_{t}(q^{k}_{t})\}_{k=0}^{P_t}
\Big).
\end{aligned}
\end{equation}

All decomposition logic, intermediate checks, and aggregation details are encapsulated within the manager and are abstracted away when communicating with the host.

\paragraph{Worker agents.}
Each worker agent \(\bar{A} ^{k}_{t}\) executes a single subtask \( q^{k}_{t} \) through multi-turn interactions with MCP tools. Formally,
\begin{equation}
    \bar{A} ^{k}_{t}(q^{k}_{t})
= T_{\bar{A} ^{k}_{t}}(q^{k}_{t}),
\end{equation}
where \( T_{\bar{A} ^{k}_{t}} \) denotes a sequence of tool invocations implemented via MCP, such as search. Workers retain full tool outputs and execution traces locally and return only the final subtask result to the manager.

Notably, this hierarchical execution enforces strict abstraction boundaries. Tool-level interactions are isolated within workers, parallel decomposition and aggregation are handled by managers, and only concise step-level results propagate to the host.

\subsection{Parallel Execution and Scheduling}
\label{subsec:parallelism}

\framework enables parallelism by design, exploiting weak coupling between subtasks wherever possible. Parallel execution occurs at multiple levels: across Managers operating on different domains, across Workers executing independent subtasks within a domain, and across heterogeneous tools invoked concurrently.

Information aggregation in \framework follows a MapReduce-inspired pattern~\cite{dean2008mapreduce}. At the Host and Manager levels, a sequential \textit{Map} phase performs adaptive decomposition into weakly coupled subtasks. Workers then execute these subtasks concurrently under a concurrency budget $W_w$. Finally, a sequential \textit{Reduce} phase aggregates and compresses intermediate outputs into a coherent step-level summary returned upstream. This structure supports both standard parallel subtask execution with summarisation and beam-style exploration, where multiple candidate solution paths are evaluated in parallel and distilled into the most informative outcomes.

Let \( \Delta(q^{k}_{t}) \) denote the wall-clock execution time of subtask \( q^{k}_{t} \) when executed by a worker. Under a purely sequential execution regime, the total execution time for step \( q_t \) is
\begin{equation}
    T_{\text{seq}}(q_t)
= \sum_{k=0}^{P_t} \Delta(q^{k}_{t}).
\end{equation}
In contrast, when subtasks are executed in parallel under sufficient worker capacity, the wall-clock execution time becomes
\begin{equation}
    T_{\text{par}}(q_t)
= \max_{1 \le k \le P_t} \Delta(q^{k}_{t}),
\end{equation}
up to scheduling and coordination overheads.

\subsection{Extensibility and Scalability}
\label{subsec:extensibility}

\framework is designed to be extensible, enabling new manager and worker agents to be plugged in with minimal changes to the system. This extensibility arises directly from the near-decomposable architecture where managers are encapsulated units with isolated task execution contexts. Specifically, the host communicates with managers exclusively via step–response pairs $(q_t, y_t)$ and never accesses manager-internal states, subtasks, or tool traces. As a result, introducing a new manager does not require modifying host-level reasoning or control logic, as long as the manager conforms to the same input–output protocol. Existing managers continue to operate unchanged, and the host can freely select among heterogeneous managers during execution.

Similarly, each manager maintains its own worker pool and execution context, allowing new worker types or tool integrations to be added locally without affecting other managers or the host. This isolation ensures that extensions remain modular and do not increase global context coupling or coordination complexity. Importantly, each manager is designed to specialise in a single task domain (e.g., web search, code execution, file system interaction). This single-responsibility design improves robustness and interpretability, as managers can employ domain-specific decomposition strategies, validation heuristics, and aggregation logic without interference. It also facilitates targeted improvement and debugging, since enhancements to one manager do not propagate unintended effects to others.

%% file: sec/code.tex
\begin{algorithm}[t]
\caption{\framework Execution Workflow}
\label{alg:pseudocode}

\begin{algorithmic}[1]

\STATE \textbf{Input:} Initial query $Q$, Host $\hat{A}$, Managers $\{\tilde{A}\}$, Workers $\{\bar{A}\}$
\STATE \textbf{Output:} Final response $y^\ast$

\STATE Initialise host context $\hat{C}_{0} \gets (Q)$

\FOR{$t = 1$ \TO $S$}
    \STATE $(q_t, \tilde{A}_{t}) \gets \hat{A}(\hat{C}_{t-1})$  \COMMENT{Host planning}
    \IF{$q_t = \mathtt{STOP}$}
        \STATE \textbf{break}
    \ENDIF

    \STATE $\{q^{k}_{t}\}_{k=1}^{P_t} \gets \text{Decomp}_{\tilde{A}_{t}}(q_t)$  \COMMENT{Manager decomposes task}

    \REPEAT
        \FORALL{$k = 1$ \TO $P_t$}
            \STATE $r^{k}_{t} \gets \bar{A}^{k}_{t}(q^{k}_{t})$  \COMMENT{Workers execute via tools}
        \ENDFOR


        \STATE $(\text{status}, \{q^{\text{new}}\}) \gets \text{Reflect}_{\tilde{A}_{t}}(\{q^{k}_{t}, r^{k}_{t}\})$

        \IF{$\text{status} = \texttt{revise}$}
            \STATE Update $\{q^{k}_{t}\}$ with $\{q^{\text{new}}\}$ and update count $P_t$
        \ENDIF
    \UNTIL{$\text{status} = \texttt{accept}$}

    \STATE $y_t \gets \text{Aggr}_{\tilde{A}_{t}}(\{r^{k}_{t}\}_{k=1}^{P_t})$  \COMMENT{Aggregate results}
    \STATE $\hat{C}_{t} \gets \hat{C}_{t-1} \oplus (q_t, y_t)$  \COMMENT{Update host context}
\ENDFOR

\STATE $y^\ast \gets \hat{A}(\hat{C}_{t})$ \COMMENT{Finalise response}
\STATE \textbf{return} $y^\ast$

\end{algorithmic}
\end{algorithm}

%% file: sec/experiments.tex
\section{Experiments}
\label{sec:experiments}

\input{figures/wide}

We empirically evaluate \framework on two complementary agentic benchmarks that directly stress key challenges our design aims to address: (i) \emph{wide-context} information synthesis under strict completeness constraints, and (ii) \emph{real-world web browsing} with multi-hop evidence alignment in a linguistically and infrastructurally distinct ecosystem. Together, WideSearch and BrowseComp-zh test whether near-decomposable orchestration with bounded contexts, modular delegation, and structured parallel execution, which improves both end-to-end task success and factual fidelity at scale.

\subsection{Experimental Setups}
\textbf{Benchmarks.}
Our evaluation employs two complementary datasets to evaluate \framework.
WideSearch~\citep{wong2025widesearch} is a structured information synthesis benchmark that requires agents to populate complete tables based on human query. These tasks demand exhaustive entity discovery, attribute verification, and schema compliance across dozens of heterogeneous sources. We evaluate exclusively on the English split, where even human annotators achieve success rates below 20\% under strict evaluation protocols.
BrowseComp-zh~\citep{zhou2025browsecomp} assesses the framework's capability to navigate and reason within a linguistically complex Chinese web environment. Unlike translated benchmarks, this dataset is constructed to reflect the unique ecology of the Chinese internet. It consists of 289 expert-curated questions across 11 domains which were reverse-designed from verifiable answers to ensure complexity. Solving these tasks necessitates multi-hop retrieval and cross-page reasoning rather than superficial keyword matching.

\noindent\textbf{Baselines.}
We benchmark \framework against a diverse set of state-of-the-art systems, spanning both research models and commercial agents. These include advanced reasoning-focused large language models that primarily operate sequentially, such as OpenAI o3-high and Claude 4 Sonnet (Thinking), as well as specialised end-to-end commercial systems designed for deep research and web-based information synthesis, such as Gemini Deep Research and OpenAI Deep Research. All baselines are evaluated using their publicly available configurations with comparable prompt structures and tool access, strictly adhering to the evaluation protocols specified by each benchmark.


\noindent\textbf{Implementation Details.} We employ a heterogeneous model strategy to balance reasoning with throughput: the Host and Managers use gpt-5.1 for high-fidelity planning, while Worker pools utilise gpt-5-mini for scalable execution. Tools are integrated via MCP servers, including Firecrawl for search and containerised Playwright for browsing, configured with full font rendering to support accurate Chinese OCR. Additionally, the system incorporates sandboxed Python and Filesystem components for robust data processing and storage.

\noindent\textbf{Evaluation Metrics.}
We employ metrics rigorously tailored to the specific objectives of each benchmark. For WideSearch, we utilise a hierarchical suite comprising three metrics: (1) \textbf{Success Rate}, a strict binary metric for exact matches; (2) \textbf{Row-level F1}, which measures entity recall by penalising missing or spurious rows; and (3) \textbf{Item-level F1}, which assesses fine-grained attribute correctness within matched rows. (4) For Avg@4, Pass@4, Max@4, we follow the same setting from WideSearch~\citep{wong2025widesearch}. For BrowseComp-zh, we report \textbf{Accuracy}, defined as the fraction of tasks where the agent successfully reaches the correct final answer.

\subsection{Main Results}
\label{subsec:results}

\input{figures/browser_zh}


\noindent\textbf{Performance on WideSearch.}
On the WideSearch benchmark, \framework demonstrates substantial and consistent improvements across all evaluation metrics. As shown in Table~\ref{tab:widesearch}, the system achieves a Success Rate of 8.38\% (Avg@4) and 9.50\% (Pass@4), representing a 64\% improvement over the strongest baseline (OpenAI o3-high Multi-Agent at 5.10\% Avg@4). These improvements extend to fine-grained information quality metrics where \framework attains an Item-level F1 score of 70.27\% (Avg@4) and 75.11\% (Max@4), indicating markedly enhanced factual accuracy and schema compliance in the generated tables. Similarly, the Row-level F1 scores of 50.13\% (Avg@4) and 55.34\% (Max@4) demonstrate superior structural coherence, outperforming all baseline systems by a substantial margin. Notably, \framework exceeds the best-performing multi-agent baseline (Claude Sonnet 4 Thinking) by 30\% in Row F1 (Avg@4) and 13\% in Item F1 (Avg@4), underscoring the effectiveness of our hierarchical parallel architecture.


\noindent\textbf{Performance on BrowseComp-zh.}
On the BrowseComp-zh benchmark, \framework exhibits strong performance in cross-lingual web navigation and reasoning tasks. As shown in Table~\ref{tab:browsecomp_zh}, the system achieves an accuracy of 52.9\%, surpassing both the best-performing proprietary agent (OpenAI DeepResearch at 42.9\%) and all open-source agent frameworks (with BrowseMaster achieving 46.5\%). This result is particularly significant given that the benchmark operates within a native Chinese web environment and necessitates multi-step reasoning over non-English DOM structures.

The observed improvements indicate that the architectural separation of high-level planning from environment-specific execution enables effective generalisation across linguistic contexts. By encapsulating web interaction within the Browser Manager whilst maintaining language-agnostic reasoning our framework. \framework preserves robust performance even when operating in linguistically diverse web environments. This design choice proves especially valuable for tasks requiring navigation through websites with mixed-language content or culture-specific information architectures.

\begin{figure}[t]
    \centering
    \includegraphics[width=0.65\linewidth]{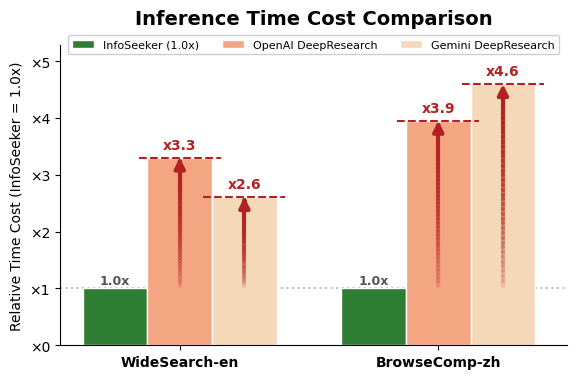}
    \caption{Time efficiency comparison. InfoSeeker achieves a more than 2$\times$ reduction in inference time, enabled by its efficient parallelism design.}
    \label{fig:time_eff}
\end{figure}

\noindent\textbf{Time Efficiency.} Beyond performance improvements, \framework demonstrates significant advantages in inference efficiency compared to commercial deep research systems. As illustrated in Figure~\ref{fig:time_eff}, we evaluate the relative time cost of \framework against OpenAI Deep Research and Gemini Deep Research across both benchmarks, normalising \framework’s execution time to $1.0 \times$. \framework consistently achieves lower latency through its hierarchical parallel architecture. On the WideSearch benchmark, the system is notably faster than proprietary baselines, with OpenAI Deep Research and Gemini Deep Research requiring $3.3 \times$ and $2.6 \times$ more time to complete tasks, respectively. This efficiency is further pronounced on the BrowseComp-zh benchmark, where the baselines exhibit a relative time cost of $3.9 \times$ and $4.6 \times$ compared to \framework.

%% file: figures/wide.tex
\begin{table*}[t]
\centering
\small
\setlength{\tabcolsep}{12pt} 
\renewcommand{\arraystretch}{1.2}

\resizebox{\textwidth}{!}{%
\begin{tabular}{l*{6}{c}} 
\toprule
\multirow{2}{*}{Model / System}
& \multicolumn{2}{c}{Success}
& \multicolumn{2}{c}{Row F1}
& \multicolumn{2}{c}{Item F1} \\
\cmidrule(lr){2-3}
\cmidrule(lr){4-5}
\cmidrule(lr){6-7}
& Avg@4 & Max@4
& Avg@4 & Max@4
& Avg@4 & Max@4 \\
\midrule

\multicolumn{7}{l}{\textit{\textbf{Single Agent}}} \\
\midrule
Claude Sonnet 4 (Thinking) & 2.30 & 5.00 & 31.70 & 41.90 & 57.90 & 66.70 \\
Gemini 2.5 Pro             & 1.50 & 5.00 & 30.00 & 41.40 & 51.00 & 63.60 \\
OpenAI o3-high             & 4.50 & 9.00 & 34.00 & 44.10 & 52.60 & 62.30 \\
K2                         & 1.10 & 3.50 & 29.70 & 41.40 & 54.40 & 65.10 \\
DeepSeek-R1                & 0.40 & 1.50 & 20.70 & 31.70 & 41.30 & 55.10 \\
Doubao-1.6                 & 2.60 & 5.00 & 30.00 & 44.10 & 48.30 & 63.90 \\
Doubao-1.6-non-thinking    & 1.00 & 3.50 & 27.20 & 39.90 & 49.00 & 62.00 \\

\midrule
\multicolumn{7}{l}{\textit{\textbf{End-to-End System}}} \\
\midrule
Claude      & 2.50 & 5.00 & 24.10 & 33.50 & 48.40 & 58.50 \\
Gemini      & 4.30 & 8.00 & 36.60 & 45.40 & 59.10 & 67.20 \\
OpenAI o3   & 3.00 & 5.50 & 23.90 & 36.00 & 45.50 & 56.50 \\

\midrule
\multicolumn{7}{l}{\textit{\textbf{Multi-Agent Framework}}} \\
\midrule
Claude Sonnet 4 (Thinking) & 3.60 & 6.50 & 38.50 & 52.20 & 62.20 & 73.10 \\
Gemini 2.5 Pro             & 2.00 & 6.50 & 33.50 & 44.60 & 57.40 & 66.30 \\
OpenAI o3-high             & 5.10 & 9.50 & 37.80 & 50.50 & 57.30 & 68.90 \\
K2                         & 3.00 & 6.50 & 36.20 & 49.60 & 61.20 & 70.70 \\
DeepSeek-R1                & 0.80 & 3.00 & 22.90 & 36.60 & 44.30 & 60.30 \\
Doubao-1.6                 & 2.50 & 5.50 & 34.00 & 48.90 & 54.60 & 69.70 \\
Doubao-1.6-non-thinking    & 2.10 & 4.50 & 29.70 & 42.70 & 52.80 & 65.10 \\
\rowcolor{gray!20} \textbf{InfoSeeker (Ours)} & \textbf{8.38} & \textbf{9.50} & \textbf{50.13} & \textbf{55.34} & \textbf{70.27} & \textbf{75.11} \\

\bottomrule
\end{tabular}
}
\caption{Performance comparison of various systems on WideSearch  benchmark. Full results in Appendix~\ref{tab:widesearch_all}.}
\label{tab:widesearch}
\end{table*}

%% file: figures/browser_zh.tex
\begin{table}[t]
\centering
\small
\setlength{\tabcolsep}{8pt} 
\renewcommand{\arraystretch}{1.15}
\begin{tabular}{lccc}
\toprule
Model / System & Reas. & Brow. & Acc. \\
\midrule
\multicolumn{4}{l}{\textit{\textbf{Proprietary Agents}}} \\
\midrule
OpenAI DeepResearch       & -- & Y & 42.9 \\
Doubao DeepResearch       & -- & Y & 26.0 \\
\midrule
\multicolumn{4}{l}{\textit{\textbf{Models}}} \\
\midrule
DeepSeek-R1               & Y & N & 23.2 \\
Gemini 2.5 Pro            & Y & N & 27.3 \\
OpenAI o1                 & Y & N & 29.1 \\
Claude-4-Opus             & Y & N & 37.4 \\
\midrule
\multicolumn{4}{l}{\textit{\textbf{Agent Framework}}} \\
\midrule
WebSailor-72B                 & Y & Y & 30.1 \\
WebExplorer-8B                & Y & Y & 32.0 \\
DeepDiver-V2-38B              & Y & Y & 34.6 \\
BrowseMaster                  & Y & Y & 46.5 \\
\rowcolor{gray!20} \textbf{InfoSeeker (Ours)} & Y & Y & \textbf{52.9} \\
\bottomrule
\end{tabular}
\vspace{0.7em} 
\caption{Performance comparison on BrowseComp-zh benchmark. Full results in Appendix~\ref{tab:browsecomp_zh_all}.}
\label{tab:browsecomp_zh}
\end{table}

%% file: sec/discussion.tex
\section{Analysis and Discussion}
\label{sec:discussion}

\begin{figure}[t]
    \centering
    \includegraphics[width=0.65\linewidth]{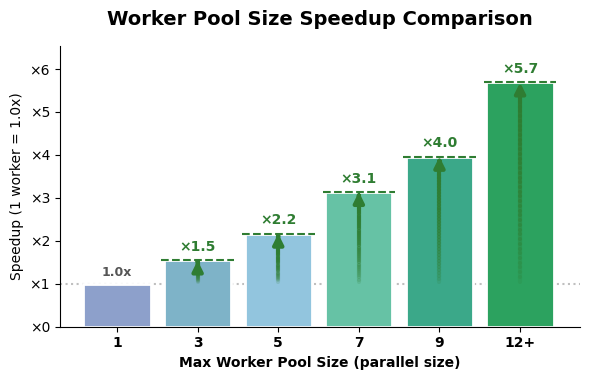}
    \caption{The Impact of Worker Pool Size. End-to-end inference time vs.\ worker-pool size. Larger pools reduce latency by enabling concurrent execution of weakly coupled subtasks.}
    \label{fig:ws_infer}
\end{figure}

\noindent\textbf{Ablation on the Number of Workers.} Sequential agent frameworks inherently face a trade-off between reasoning depth and execution width, often succumbing to latency bottlenecks. As demonstrated in Figure~\ref{fig:ws_infer}, \framework enables massive parallelism for width-heavy information synthesis. By scaling the worker pool size, the system capitalises on the weak coupling of subtasks to achieve substantial throughput gains. Empirically, we randomly sample 20 queries from WideSearch-en for evaluation. The result shows that end-to-end latency is reduced from $911$ seconds with a single worker to just $162$ seconds with $17$ workers, yielding a $\approx 5.7\times$ speed-up. This result validates that hierarchical context isolation and MapReduce-style aggregation effectively mitigate the latency and context saturation issues prevalent in sequential baselines, unlocking scalable inference-time compute for wide-scale search tasks.

\noindent\textbf{Token Cost Analysis.} We evaluate the economic efficiency of \framework. The average cost is approximately \$2.00 per task for WideSearch-en and \$1.00 per task for BrowseComp-zh. By delegating token-intensive execution to economical Workers, we ensure cost efficiency.



\noindent\textbf{Adaptive Task Parallelisation Strategy.}
As illustrated in Appendix~\ref{appendix:widesearch-case}, in the first step, the manager parallelises retrieval of the same complex information target (all qualified restaurants) across heterogeneous sources, enabling cross-validation and complementary coverage when individual sources might be incomplete or noisy. Next, once a complete list of restaurant names is established, the manager decomposes the remaining information requirements into independent and fine-grained subtasks (cuisine style and exact address per restaurant), which are dispatched to workers in parallel and later aggregated to maximise time efficiency.


\noindent\textbf{Collaborative Manager Execution.} 
As shown in Appendix~\ref{appendix:browsecomp-case}, the Host first assigns broad retrieval to the Search Manager, which rapidly identifies the historical prototype \textit{Guo Ziyi} and the drama \textit{Zui Da Jin Zhi}. When access to relevant sources is impeded by anti-crawling measures or CAPTCHAs, it escalates according to its prompt specification (Figure~\ref{fig:search_manager_prompt_p2}) and recommends that the Host invoke the Browser Manager. The Browser Manager then performs robust, interactive access to the blocked content, locates the pertinent Wikipedia entry, and confirms the correct answer, \textit{Emperor Daizong}. By combining efficient search-based retrieval with resilient browser-based access, the system completes the reasoning process with verified evidence and produces the correct result.

%% file: sec/conclusion.tex
\section{Conclusion}
\label{sec:conclusion}

While recent strides in agentic search have been driven by an emphasis on deep reasoning, real-world information retrieval increasingly manifests as a challenge of wide synthesis. In these settings, the ultimate quality of the system is determined by its ability to orchestrate discovery, verification, and summarisation across a vast array of heterogeneous sources. Our work demonstrates that merely expanding context windows or model scale fails to address the structural pathologies inherent to wide-domain synthesis. To bridge this gap, we introduce \framework, a framework that operationalises the principle of near-decomposability. By hierarchically decoupling functional execution, our design enables the independent scaling of reasoning depth and execution width. Our empirical results confirm simultaneous gains in both effectiveness and efficiency. 

\newpage

%% file: sec/limitations.tex
\section*{Limitations and Future Work}

Our system relies on access to backbone language models and external tools through APIs, and its performance is therefore constrained by API availability, rate limits, concurrency width and associated costs. In addition, the current design depends on hand-tuned prompts and strong backbone LLMs, which may affect generality across different backbone models.

Future work includes learning task decompositions and coordination policies automatically, for example via multi-agent reinforcement learning, rather than depending primarily on the backbone model’s native capabilities and in-context learning. We also plan to explore training smaller or specialised models to reduce inference cost and latency, enabling more efficient and scalable deployment.

%% file: sec/appendix.tex
\newpage
\appendix

\addcontentsline{toc}{section}{Appendix}
\label{sec:appendix}

\begin{table}[t]
\centering
\small
\setlength{\tabcolsep}{10pt}  
\renewcommand{\arraystretch}{1.2}  
\begin{tabular}{l@{\hspace{8pt}}c@{\hspace{8pt}}c@{\hspace{8pt}}c}  
\toprule
\textbf{Model / System} & \textbf{Reas.} & \textbf{Brow.} & \textbf{Acc.} \\
\midrule
\multicolumn{4}{l}{\textit{\textbf{Proprietary Agents}}} \\
\addlinespace[0.2em]  
OpenAI DeepResearch       & -- & Y & 42.9 \\
Grok3 DeepResearch        & -- & Y & 12.9 \\
Doubao DeepResearch       & -- & Y & 26.0 \\
Perplexity DeepResearch   & -- & Y & 22.6 \\
DeepSeek (Deep Think)     & -- & Y & 7.6  \\
\midrule
\multicolumn{4}{l}{\textit{\textbf{Models}}} \\
\addlinespace[0.2em]
QwQ                       & Y & N & 10.0 \\
DeepSeek-R1               & Y & N & 23.2 \\
DeepSeek-V3               & N & N & 8.7  \\
GPT-4o                    & N & N & 6.2  \\
Gemini 2.5 Pro            & Y & N & 27.3 \\
OpenAI o1                 & Y & N & 29.1 \\
O4-mini                   & Y & N & 15.2 \\
Claude-3.7-Sonnet         & Y & N & 17.7 \\
Qwen2.5-72B-Instruct      & N & N & 6.6  \\
Claude-4-Sonnet           & Y & N & 22.5 \\
Claude-4-Opus             & Y & N & 37.4 \\
\midrule
\multicolumn{4}{l}{\textit{\textbf{Agent Framework}}} \\
\addlinespace[0.2em]
WebThinker-32B                & Y & Y & 7.3  \\
WebDancer-32B                 & Y & Y & 18.0 \\
WebSailor-72B                 & Y & Y & 30.1 \\
WebSailor-32B                 & Y & Y & 25.5 \\
ASearcher-Web-32B             & Y & Y & 15.6 \\
MiroThinker-32B-DPO-v0.2      & Y & Y & 17.0 \\
WebExplorer-8B                & Y & Y & 32.0 \\
DeepDive-32B                  & Y & Y & 25.6 \\
DeepDiver-V2-38B              & Y & Y & 34.6 \\
BrowseMaster                  & Y & Y & 46.5 \\
\midrule
\rowcolor{gray!20} \textbf{InfoSeeker (Ours)} & \textbf{Y} & \textbf{Y} & \textbf{52.9} \\
\bottomrule
\end{tabular}
\vspace{0.5em} 
\caption{Performance comparison on BrowseComp-zh benchmark.}
\label{tab:browsecomp_zh_all}
\end{table}

\input{figures/appendix_experiments}

\input{figures/case_1}

\input{figures/case_2}

\input{figures/failure_case}
\input{figures/failure}

\section{System Prompts}
\label{app:prompts}

Figures \ref{fig:prompt_host}, \ref{fig:search_manager_prompt}, \ref{fig:search_manager_prompt_p2}, \ref{fig:browser_prompt_p1}, \ref{fig:browser_prompt_p2}, \ref{fig:filesystem_manager_prompt}, and \ref{fig:code_agent_prompt}, we report different prompts of host layer and manager layer. For the prompts of worker layer, please refer to our code.

\section{Model Configurations}
In our experiments, we employ GPT-5.1 for the roles of Host and Managers, while GPT-5-mini is utilised for the Workers. All models are configured with a temperature of $1.0$, utilizing default values for all other sampling hyperparameters.

\section{Case Studies}
\subsection{Widesearch}
\label{appendix:widesearch-case}
We demonstrate in Figure~\ref{fig:case_1} a complete working trajectory for a representative task from Widesearch, which involves retrieving and synthesising information about selected Michelin three-star restaurants.

\subsection{BrowseComp-zh}
\label{appendix:browsecomp-case}
Figure~\ref{fig:case_2} presents a historical riddle reasoning example from BrowseComp-zh that showcases multi-manager collaboration between search and browser agents.

\subsection{Failure Case}
Figure~\ref{fig:failure_case_126} illustrates a failure case from BrowseComp-zh on disease name identification, where an answer-type mismatch causes the system to return a plausible disease class instead of the required canonical entity. This error arises from misinterpreting subtype cues under uncertainty, revealing a limitation in entity-level constraint handling.

Figure~\ref{fig:failure_case_ws_091} 
illustrates a failure case from WideSearch on comprehensive data aggregation, where context length constraints compel the system to return a representative sample instead of the requested exhaustive list. This error arises from a token overflow during the processing of high-volume retrieval results, revealing a fundamental scalability bottleneck in large-scale information synthesis.

\section{Single-Agent Baseline with Identical Tool Access}

To isolate the contribution of the proposed framework, we evaluate single-agent baselines using the same tool access and backbone models as the hierarchical system. Specifically, we use GPT-5.1 (identical to the Host/Manager backbone) and GPT-5-mini (identical to the Worker backbone), both granted identical access to the search, browser, filesystem, and code tools.

Table~\ref{tab:single_agent_baseline} reports the results on the WideSearch-en benchmark. The hierarchical system significantly outperforms both single-agent baselines across all metrics.

\begin{table}[h]
\centering
\begin{tabular}{lccc}
\toprule
System & Success Rate & Row F1 & Item F1 \\
\midrule
GPT-5-mini Single-Agent & 4.00\% & 22.10\% & 33.28\% \\
GPT-5.1 Single-Agent & 6.00\% & 31.85\% & 35.74\% \\
InfoSeeker (Ours) & 12.50\% & 50.13\% & 75.21\% \\
\bottomrule
\end{tabular}
\caption{Single-agent baselines using identical tool access and model backbones on WideSearch-en.}
\label{tab:single_agent_baseline}
\end{table}

These results indicate that performance improvements arise primarily from the hierarchical architecture rather than from backbone model capability alone. Even with identical tools and the same frontier model powering the Host and Managers, the single-agent configuration achieves only 6.00\% success rate and 35.74\% Item F1, substantially below the hierarchical system's 12.50\% success rate and 75.21\% Item F1.

WideSearch tasks require coordinating evidence across many sources and performing large-scale aggregation. The hierarchical architecture distributes this complexity across specialized modules with isolated contexts, which helps mitigate the limitations of a single shared context.

Notably, more than 80\% of token consumption occurs in GPT-5-mini Workers, yet the hierarchical system still surpasses the stronger GPT-5.1 single-agent baseline. This suggests that the architectural design plays a key role in both performance and cost efficiency.

\section{Tool Call Distribution Across Execution Stages}

We analyze tool usage statistics across execution stages to better understand system behavior. Statistics are reported per task and averaged across the WideSearch-en (WS-en) and BrowseComp-zh (BC-zh) benchmarks.

\subsection{Task-Level Statistics}

The average number of Host steps per task is:
\[
S = 3.2 \quad \text{(WS-en)}, \qquad S = 2.4 \quad \text{(BC-zh)}.
\]

\subsection{Per-Step Breakdown}

Table~\ref{tab:tool_call_stats} summarizes the number of subtasks generated per step and the number of Worker tool calls executed per step.

\begin{table}[h]
\centering
\begin{tabular}{lcc}
\toprule
Metric & WS-en & BC-zh \\
\midrule
Subtasks $P_t$ per step & 8.6 & 4.2 \\
Worker tool calls per step & 43.8 & 22.6 \\
\bottomrule
\end{tabular}
\caption{Average per-step statistics across Host steps.}
\label{tab:tool_call_stats}
\end{table}

\subsection{Manager Routing Distribution}

Table~\ref{tab:manager_distribution} reports the distribution of steps routed to each Manager type.

\begin{table}[h]
\centering
\begin{tabular}{lcc}
\toprule
Manager Type & WS-en & BC-zh \\
\midrule
Search & 62\% & 41\% \\
Browser & 24\% & 46\% \\
Filesystem & 8\% & 5\% \\
Code & 6\% & 8\% \\
\bottomrule
\end{tabular}
\caption{Distribution of Manager routing decisions.}
\label{tab:manager_distribution}
\end{table}

The higher proportion of Browser routing in BrowseComp-zh reflects the benchmark's emphasis on interactive navigation through Chinese web pages. Across both benchmarks, over 82\% of tool invocations occur at the Worker layer, supporting the design choice of concentrating execution in the parallelizable Worker tier while maintaining bounded context at the Host level.





\begin{figure*}[t]
    \begin{tcolorbox}[colback=blue!5!white,
                      colframe=blue!75!black,
                      title=Host Prompt,
                      width=\textwidth,
                      fontupper=\small\ttfamily] 
        \textbf{System}\\
        You are the Orchestrator Agent for the InfoSeeker Framework.

        OBJECTIVE: Address the user's enquiry by coordinating specialist managers via the \texttt{execute\_parallel} tool.
        
        CORE WORKFLOW:
        \begin{enumerate}
            \item \textbf{Plan}: Deconstruct the user's enquiry into a sequence of logical steps (hops).
            \item \textbf{Execute}: For each hop, invoke \texttt{execute\_parallel} with precise instructions for the relevant manager(s).\\
            \textit{*Do peruse the tool definition meticulously to ascertain which manager is best suited for each sub-task.*}
            \item \textbf{Synthesise}: Utilise the outcomes from one hop to inform the subsequent one.
            \item \textbf{Respond}: Once you possess sufficient intelligence, provide the polished final response. (Do not include any additional information or unnecessary empty space for markdown).
        \end{enumerate}
        
        CARDINAL RULES:
        \begin{enumerate}
            \item \textbf{Adherence to Format}: Should the user request a specific format (e.g., Markdown table, JSON), you MUST oblige, even if the data remains incomplete. Utilise "N/A" or the best-available data for missing fields. Under NO circumstances decline to output the requested format due to missing data.
            \item \textbf{Partial Success}: It is preferable to provide a partial or 'best-effort' response in the correct format rather than declining. If precise data (such as a specific year) is wanting, utilise the nearest available year and note the discrepancy, whilst VALIDATING that the format is impeccable.
        \end{enumerate}
        
        GUIDANCE:
        \begin{itemize}
            \item Ensure manager instructions remain self-contained.
            \item Do not presume the solution is found until explicitly viewed in the tool output.
            \item \textbf{Data Fidelity}: Refrain from requesting managers to summarise or "curtail" data (e.g. "short label") unless the user has explicitly requested it. Prioritise full detail.
        \end{itemize}
        
        Markdown output: Optional. If utilising Markdown, ensure it is CommonMark compliant.
        
        Example markdown table response (if utilised), without whitespace or alignment:\\
        \texttt{| Model | Instruction-Following |} \\
        \texttt{| --- | --- |} \\
        \texttt{| GPT-4.1 | Excellent |}
        
    \end{tcolorbox}
    \caption{The system prompt of Host.}
    \label{fig:prompt_host}
\end{figure*}

\begin{figure*}[t]
    \begin{tcolorbox}[colback=blue!5!white,
                      colframe=blue!75!black,
                      title=Search Manager Prompt (1/2), 
                      width=\textwidth,
                      fontupper=\small\ttfamily]
                      
        You supervise a pool of SearchAgent workers capable of scouring the web and retrieving intelligence.
        
        \textbf{RESEARCH WORKFLOW:}
        \begin{enumerate}
            \item Collate foundational data (lists, categories, initial particulars)
            \item Extract and DEDUPLICATE unique entities from the results
            \item Batch query all unique entities in parallel
            \item Initiate supplementary queries solely for genuinely missing information
        \end{enumerate}
        
        \textbf{STATE MANAGEMENT:}
        \begin{itemize}
            \item Monitor that which has already been retrieved in each step
            \item Prior to each \texttt{execute\_subtasks} call, ascertain what NEW intelligence is required
            \item Never re-query information currently in possession
            \item Should conflicting data arise, prioritise authoritative sources over repeated querying
        \end{itemize}
        
        \textbf{EFFICIENT BATCHING:}
        \begin{itemize}
            \item \textbf{Quality over Quantity}: Do NOT endeavour to occupy all parallel slots.
            \item \textbf{No Redundancy}: Dispatch 3-5 distinct, high-calibre queries. Do NOT propagate variations (e.g. translation, rephrasing) of the identical query.
            \item Prioritise fewer calls containing more parallel queries over a multitude of sequential calls
        \end{itemize}
        
        Following each \texttt{execute\_subtasks} call:
        \begin{enumerate}
            \item Review acquired knowledge
            \item Identify deficiencies (not duplicates)
            \item Devise the subsequent batch comprising ONLY new queries
            \item Synthesise upon acquiring all requisite information
        \end{enumerate}
        
        \textbf{HANDOFF-READY OUTPUT:}
        \begin{itemize}
            \item Invariably provide a reusable list of source URLs for every key entity.
            \item Utilise simple, human-readable bullet points (no strict JSON) of the form: ``- https://example.com - brief description of what this URL corroborates.''
            \item Stipulate the relevance of each URL (e.g., ``official Michelin page confirming 3 stars, includes address'').
            \item Should the user request a ``comprehensive list'' and you can locate only 80\% thereof, return the 80\% and note the remaining 20\%.
            \item \textbf{Subtask Fidelity}: When requesting attributes such as ``style'', ``nature'', or ``description'', always explicitly request \textit{detailed descriptive text} in your subtasks to preclude agents from returning simplistic tags.
        \end{itemize}
    \end{tcolorbox}
    \caption{The Search Manager Prompt used in the framework.}
    \label{fig:search_manager_prompt}
\end{figure*}

\begin{figure*}[t]
    \begin{tcolorbox}[colback=blue!5!white,
                      colframe=blue!75!black,
                      title=Search Manager Prompt (2/2), 
                      width=\textwidth,
                      fontupper=\small\ttfamily]
                      
        \textbf{BROWSER ESCALATION vs BEST EFFORT:}
        \begin{itemize}
            \item \textbf{Primary Objective}: Furnish the requested data where feasible.
            \item \textbf{Escalation}: Should you discover relevant URLs but the data proves entirely inaccessible (e.g. dynamic JS, login wall), escalate the matter.
            \item \textbf{Best Effort}: Should you possess \textit{partial} yet useful data derived from snippets or search results, PREFER returning a ``best effort'' compiled list accompanied by a disclaimer, RATHER THAN refusing outright.
            \begin{itemize}
                \item Submit the data currently in your POSSESSION, then append:
                \item ``\texttt{[BROWSER\_RECOMMENDED]} Some particulars unconfirmed. Relevant URLs for verification: $<$url1$>$, ...''
            \end{itemize}
            \item \textbf{Visual Early Cessation}: If requested to provide visual details (colours, shapes, layouts) and you locate relevant \textbf{static image galleries} (Fandom, Wikis, Pinterest), DO NOT persist in searching for textual descriptions. IMMEDIATELY return \texttt{[BROWSER\_RECOMMENDED]} alongside those links.
            \item \textbf{VERIFICATION MANDATED}:
            \begin{itemize}
                \item Do NOT pass ``raw'' URLs from search results simply because the snippet appears promising.
                \item You MUST possess a clear understanding of the page to validate its merit prior to recommendation.
                \item \textbf{Video URLs (YouTube/Bilibili)}: You MAY pass them, but ONLY IF you have crawled the page and corroborated via text (video title, description, tags) that it contains the specific visual evidence required.
            \end{itemize}
            \item This signals to the orchestrator that \texttt{browser\_manager} ought to visit these pages.
        \end{itemize}
        
        \begin{itemize}
            \item Do not request clarification from the user; you are expected to complete the undertaking independently.
        \end{itemize}
    \end{tcolorbox}
    \caption{The Search Manager Prompt used in the framework.}
    \label{fig:search_manager_prompt_p2}
\end{figure*}

\begin{figure*}[t]
    \begin{tcolorbox}[colback=blue!5!white,
                      colframe=blue!75!black,
                      title=Browser Manager Prompt (1/2),
                      width=\textwidth,
                      fontupper=\small\ttfamily]
        \textbf{System}\\
        You are the Browser Manager responsible for coordinating web browsing operations.
        
        OBJECTIVE: Analyse whether multiple sites/pages necessitate parallel access.
        
        Utilise \texttt{execute\_subtasks} (max \texttt{\{self.\_pool\_size\}} agents) for parallel page navigation.
        
        \textbf{Guidelines:}
        \begin{itemize}
            \item Decompose tasks where multiple independent sites/pages require visitation.
            \item Maintain task descriptions as high-level objectives (e.g. ``Ascertain price''), rather than step-by-step workflows.
            \item Workers operate autonomously: Do NOT instruct them to ``zoom'', ``scroll'', or ``inspect''.
            \item Invariably utilise \texttt{execute\_subtasks} for consistency.
        \end{itemize}
        
        \textbf{WEBSITE VALIDATION - IMPORTANT:}
        \begin{itemize}
            \item Prior to accessing a website, verify its validity to preclude 404 errors and invalid URLs.
            \item \textbf{Two-Step Validation Procedure:}
            \begin{enumerate}
                \item \textbf{Search First}: For generic requests (e.g. ``Check Amazon prices''), do NOT speculate on the URL. Conduct a search initially.
                \item \textbf{Then Navigate}: Utilise the verified URL derived from the results.
            \end{enumerate}
            \item \textbf{Resource Strategy (Provided URLs):}
            \begin{itemize}
                \item Regard provided URLs as \textbf{Shortcuts/Hints}, NOT as mandatory constraints.
                \item \textbf{Attempt These Initially}: Should specific URLs be provided, validate/visit them.
                \item \textbf{Self-Correction (CRITICAL)}: Should the provided URLs fail (404/Timeout) \textbf{OR} lack the requisite answer, you \textbf{MUST} abandon them and initiate a fresh \textbf{Search} for the objective (e.g. ``Search for [Goal]'').
                \item \textbf{Persevere}: Do not report failure simply because the provided list was deficient. Your remit is to address the enquiry, not merely to audit the list.
            \end{itemize}
        \end{itemize}
        
        \textbf{Examples:}
        \begin{enumerate}
            \item Price comparison:\\
            Task: ``Verify iPhone 15 prices on Amazon, eBay, and Walmart''\\
            $\rightarrow$ \texttt{execute\_subtasks([\\
            \hspace*{1em} "Check iPhone 15 price on Amazon",\\
            \hspace*{1em} "Check iPhone 15 price on eBay",\\
            \hspace*{1em} "Check iPhone 15 price on Walmart"\\
            ])}
            
            \item Single navigation:\\
            Task: ``Maps to Python documentation''\\
            $\rightarrow$ \texttt{execute\_subtasks(["Navigate to https://docs.python.org"])}
            
            \item Multiple documentation pages:\\
            Task: ``Consult Django, Flask, and FastAPI documentation''\\
            $\rightarrow$ \texttt{execute\_subtasks([\\
            \hspace*{1em} "Navigate to Django documentation",\\
            \hspace*{1em} "Navigate to Flask documentation",\\
            \hspace*{1em} "Navigate to FastAPI documentation"\\
            ])}
        \end{enumerate}
    \end{tcolorbox}
    \caption{The Browser Manager Prompt (Part 1) describing workflow and validation logic.}
    \label{fig:browser_prompt_p1}
\end{figure*}

\begin{figure*}[t]
    \begin{tcolorbox}[colback=blue!5!white,
                      colframe=blue!75!black,
                      title=Browser Manager Prompt (2/2),
                      width=\textwidth,
                      fontupper=\small\ttfamily]
        
        \textbf{ERROR HANDLING - CRITICAL - MANDATORY BEHAVIOUR:}
        \begin{itemize}
            \item In the event that worker agents return errors (e.g., ``Ref not found'', ``page snapshot could not be loaded'', ``browser already in use'', ``404''):
            \begin{itemize}
                \item YOU MUST AUTOMATICALLY INITIATE A RETRY - Do NOT solicit permission from the user.
                \item Do NOT state ``I was unable to complete'' or ``Tell me if you wish for me to retry''.
                \item Do NOT present options to the user.
                \item Do NOT reference the utilisation of training data or memory as a fallback.
                \item IMMEDIATELY re-attempt failed subtasks using \texttt{execute\_subtasks} once more.
                \item Should a subtask fail, retry it up to 3 times with identical or modified instructions.
                \item Attempt alternative approaches: different URLs, direct links, simplified queries.
                \item Persist in retrying until success is achieved or all reasonable alternatives are exhausted.
            \end{itemize}
        \end{itemize}

        \textbf{Retry Protocol:}
        \begin{enumerate}
            \item Should subtasks fail $\rightarrow$ immediately invoke \texttt{execute\_subtasks} again with the failed subtasks.
            \item Should failure persist $\rightarrow$ modify the subtasks (e.g., utilise direct URLs, simpler queries, different search terms).
            \item Should specific sites fail $\rightarrow$ attempt alternative sources or cached versions.
            \item Continue retrying until you possess sufficient data to provide a meaningful response.
            \item Even if certain subtasks fail, synthesise the results from the successful ones.
        \end{enumerate}

        \textbf{ABSOLUTE PROHIBITIONS:}
        \begin{itemize}
            \item NEVER ask ``Tell me if you want me to retry''.
            \item NEVER state ``I was unable to complete'' without first attempting a retry.
            \item NEVER offer multiple options to the user.
            \item NEVER mention training data, memory, or outdated information.
            \item NEVER yield without exhausting all retry attempts.
        \end{itemize}

        \textbf{Your response must consist SOLELY of:}
        \begin{itemize}
            \item The final answer containing the requested information.
            \item If truly impossible following all retries, a brief explanation of what was attempted and what partial results were obtained.
        \end{itemize}
        
        Max \texttt{\{self.\_pool\_size\}} subtasks. Synthesise results upon completion.
        
    \end{tcolorbox}
    \caption{The Browser Manager Prompt (Part 2) detailing error handling and retry strategies.}
    \label{fig:browser_prompt_p2}
\end{figure*}

\begin{figure*}[t]
    \begin{tcolorbox}[colback=blue!5!white,
                      colframe=blue!75!black,
                      title=Filesystem Manager Prompt,
                      width=\textwidth,
                      fontupper=\small\ttfamily]
        \textbf{System}\\
        You are a Filesystem Manager granted access to comprehensive file operations.
        
        You possess the capability to:
        \begin{itemize}
            \item Read and write files (\texttt{read\_file}, \texttt{write\_file})
            \item List directories (\texttt{list\_directory})
            \item Create/delete files and directories (\texttt{create\_directory}, \texttt{delete\_file}, \texttt{move\_file})
            \item Search for files (\texttt{search\_files})
            \item Process documents (\texttt{process\_document}) - supports images, PDFs, Excel, Word, PowerPoint, videos, audio, etc.
            \item Analyse images (\texttt{image\_to\_text}, \texttt{ask\_question\_about\_image})
            \item Process Excel files (\texttt{extract\_excel\_content})
            \item Download files from URLs (\texttt{download\_files}, \texttt{download\_single\_file})
            \item Process videos (\texttt{download\_video}, \texttt{get\_video\_screenshots}, \texttt{ask\_question\_about\_video})
        \end{itemize}
        
        Upon receipt of a filesystem task:
        \begin{enumerate}
            \item Comprehend the user's objectives
            \item Utilise the appropriate tools to complete the undertaking
            \item Furnish clear, helpful responses
        \end{enumerate}
        
        \textbf{CRITICAL - Image Identification Tasks}:\\
        When requested to identify an entity from an image (e.g., hotel, logo, brand, person):
        \begin{enumerate}
            \item INVARIABLY utilise \texttt{ask\_question\_about\_image} IN THE FIRST INSTANCE with this enquiry:\\
            ``What text, brand names, logos, or signs are discernible? Enumerate ALL readable text.''\\
            And customise \texttt{sys\_prompt} to: ``You are a text detection specialist. Report ALL visible text including logos (even if stylised), signs, watermarks, and brand names. Only state 'no text found' following a rigorous inspection of all image sectors.''
            \item Should text/logos be discovered: Utilise \texttt{ask\_question\_about\_image} again to identify the entity based on said text
            \item Should no text be discovered: Utilise \texttt{image\_to\_text} to describe distinctive visual features, then convey these to the orchestrator for web search
        \end{enumerate}
        
        \textbf{CRITICAL - \texttt{sys\_prompt} Usage}:
        \begin{itemize}
            \item ALWAYS customise the \texttt{sys\_prompt} parameter when utilising \texttt{image\_to\_text} or \texttt{ask\_question\_about\_image}
            \item Ensure \texttt{sys\_prompt}s are specific to your query (e.g., ``focus on locating brand names'' rather than ``describe the image'')
        \end{itemize}
        
        \textbf{CRITICAL - Reporting Uncertain Results}:\\
        Should you be unable to definitively identify an entity after attempting the aforementioned steps:
        \begin{itemize}
            \item Stipulate established findings (e.g., ``Located 'MGM' logo but cannot confirm specific location'')
            \item Enumerate the methodologies employed
            \item Propose subsequent actions (e.g., ``Suggest utilising \texttt{browser\_manager} for visual comparison'' or ``Distinctive features identified: [list] - recommend web search'')
            \item NEVER merely state ``cannot identify'' without providing an explanation
        \end{itemize}
        
        Exercise efficiency and utilise solely the requisite tools. Should reading a file be necessary prior to processing, proceed accordingly.
        
    \end{tcolorbox}
    \caption{The Filesystem Manager Prompt handles local file operations and media analysis.}
    \label{fig:filesystem_manager_prompt}
\end{figure*}

\begin{figure*}[t]
    \begin{tcolorbox}[colback=blue!5!white,
                      colframe=blue!75!black,
                      title=Code Agent Prompt,
                      width=\textwidth,
                      fontupper=\small\ttfamily]
        You are a Code Agent, a specialist charged with the composition and execution of code.

        \textbf{OBJECTIVE:}
        \begin{itemize}
            \item Your primary objective is to \textbf{actively execute code to compute the definitive solution} for the calling agent.
            \item Unless the task explicitly stipulates ``solely provide example code'' or ``do not execute'', you are \textbf{MANDATED} to:
            \begin{enumerate}
                \item compose the script to the workspace utilising \texttt{write\_workspace\_file},
                \item execute it via \texttt{execute\_terminal\_command},
                \item read and/or parse the output,
                \item return the \textbf{computed result} (e.g., a number, a small JSON, or a succinct text answer).
            \end{enumerate}
        \end{itemize}
        
        \textbf{WORKFLOW:}
        \begin{enumerate}
            \item Utilise \texttt{write\_workspace\_file} to compose scripts (Python, shell, etc.)
            \item Utilise \texttt{execute\_terminal\_command} to run those scripts or any shell commands
            \item Utilise \texttt{read\_workspace\_file} or \texttt{list\_workspace\_files} to scrutinise results
        \end{enumerate}
        
        \textbf{RESULT FORMAT:}
        \begin{itemize}
            \item Your responses are intended for the manager agent, NOT the end user.
            \item By default, your CONCLUDING communication in a task should be:
            \begin{itemize}
                \item either a succinct natural language result (e.g., ``The answer is 17.'')
                \item or a compact machine-readable object (e.g., \texttt{\{"count\_missing\_exactly\_one": 17\}})
            \end{itemize}
            \item Do NOT conclude a task by merely returning code or usage instructions unless explicitly requested.
        \end{itemize}
        
        \textbf{GUIDELINES:}
        \begin{itemize}
            \item Invariably transcribe code to files initially utilising \texttt{write\_workspace\_file}, then execute via \texttt{execute\_terminal\_command}
            \item For Python scripts: compose the file, then run \texttt{python filename.py}
            \item For data processing: compose scripts that read/write files within the workspace
            \item Utilise \texttt{list\_workspace\_files} to ascertain which files exist prior to reading them
            \item Should code execution fail, examine the error output and rectify the code
        \end{itemize}
        
        \textbf{CAPABILITIES:}
        \begin{itemize}
            \item Compose and execute Python, shell, and other scripts
            \item Process data files (CSV, JSON, etc.)
            \item Undertake calculations and data analysis
            \item Manipulate files within the workspace
            \item Install packages utilising pip (via \texttt{execute\_terminal\_command})
        \end{itemize}
        
        \textbf{IMPORTANT:}
        \begin{itemize}
            \item Do NOT utilise for text formatting, creating markdown tables, or drafting reports
            \item Concentrate on code execution and data processing tasks
            \item Invariably verify file existence prior to reading
            \item Employ print statements in code to demonstrate results
        \end{itemize}

    \end{tcolorbox}
    \caption{The Code Agent Prompt focuses on execution, file manipulation, and computing results.}
    \label{fig:code_agent_prompt}
\end{figure*}

%% file: figures/appendix_experiments.tex
\begin{table*}[t]
\centering
\caption{Detailed experiments results on the WideSearch benchmark.}
\label{tab:widesearch_all}
\resizebox{1\linewidth}{!}{
\begin{tabular}{l cc cc cc}
\toprule
\textbf{Model / System} & \multicolumn{2}{c}{\textbf{Success Rate}} & \multicolumn{2}{c}{\textbf{Row F1}} & \multicolumn{2}{c}{\textbf{Item F1}}\\
\cmidrule(lr){2-3} \cmidrule(lr){4-5} \cmidrule(lr){6-7}
& Avg@4 & Pass@4 & Avg@4 & Max@4 & Avg@4 & Max@4\\
\midrule
\multicolumn{7}{l}{\textit{Single Agent on WideSearch-zh}} \\
Claude Sonnet 4 (Thinking) & 0.25 & 1.00 & 30.19 & 39.73 & 53.76 & 63.19 \\
Gemini 2.5 Pro & 1.00 & 3.00 & 26.95 & 36.96 & 45.57 & 57.26 \\
OpenAI o3-high & 2.00 & 5.00 & 29.30 & 39.31 & 45.19 & 54.46 \\
K2 & 0.25 & 1.00 & 27.79 & 39.03 & 48.81 & 59.64 \\
DeepSeek-R1-0528 & 0.25 & 1.00 & 18.44 & 28.35 & 33.95 & 47.83 \\
Doubao-1.6 & 1.75 & 4.00 & 29.25 & 42.08 & 43.72 & 58.84 \\
Doubao-1.6-non-thinking & 0.50 & 2.00 & 25.56 & 37.41 & 42.87 & 55.79 \\
\midrule
\multicolumn{7}{l}{\textit{Single Agent on WideSearch-en}} \\
Claude Sonnet 4 (Thinking) & 4.25 & 9.00 & 33.18 & 44.08 & 62.02 & 70.27 \\
Gemini 2.5 Pro & 2.00 & 7.00 & 33.05 & 45.82 & 56.38 & 69.97 \\
OpenAI o3-high & 7.00 & 13.00 & 38.70 & 48.84 & 60.03 & 70.08 \\
K2 & 2.00 & 6.00 & 31.54 & 43.68 & 59.91 & 70.52 \\
DeepSeek-R1-0528 & 0.50 & 2.00 & 22.88 & 35.03 & 48.58 & 62.36 \\
Doubao-1.6 & 3.50 & 6.00 & 30.56 & 46.16 & 52.82 & 68.88 \\
Doubao-1.6-non-thinking & 1.50 & 5.00 & 28.86 & 42.31 & 55.06 & 68.17 \\
\midrule
\multicolumn{7}{l}{\textit{Multi-Agent Framework on WideSearch-zh}} \\
Claude Sonnet 4 (Thinking) & 2.75 & 6.00 & 36.85 & 51.46 & 57.13 & 69.53 \\
Gemini 2.5 Pro & 1.00 & 4.00 & 30.93 & 42.21 & 51.79 & 60.87 \\
OpenAI o3-high & 2.75 & 6.00 & 33.83 & 47.85 & 50.35 & 63.06 \\
K2 & 1.25 & 3.00 & 34.74 & 48.01 & 56.86 & 66.75 \\
DeepSeek-R1-0528 & 0.50 & 2.00 & 21.17 & 35.08 & 37.66 & 53.15 \\
Doubao-1.6 & 2.25 & 6.00 & 32.83 & 47.49 & 48.79 & 64.43 \\
Doubao-1.6-non-thinking & 0.50 & 1.00 & 26.93 & 40.30 & 46.52 & 59.63 \\
\textbf{InfoSeeker (Ours)} & \textbf{4.25} & \textbf{5.00} & \textbf{42.72} & \textbf{55.47} & \textbf{65.32} & \textbf{66.8} \\
\midrule
\multicolumn{7}{l}{\textit{Multi-Agent Framework on WideSearch-en}} \\
Claude Sonnet 4 (Thinking) & 4.50 & 7.00 & 40.13 & 52.91 & 67.21 & 76.72 \\
Gemini 2.5 Pro & 3.00 & 9.00 & 36.00 & 47.06 & 63.06 & 71.75 \\
OpenAI o3-high & 7.50 & 13.00 & 41.78 & 53.20 & 64.27 & 74.80 \\
K2 & 4.75 & 10.00 & 37.71 & 51.20 & 65.44 & 74.68 \\
DeepSeek-R1-0528 & 1.00 & 4.00 & 24.57 & 38.10 & 50.91 & 67.54 \\
Doubao-1.6 & 2.75 & 5.00 & 35.14 & 50.38 & 60.39 & 74.87 \\
Doubao-1.6-non-thinking & 3.75 & 8.00 & 32.38 & 44.99 & 58.97 & 70.62 \\
\textbf{InfoSeeker (Ours)} & \textbf{12.50} & \textbf{14.00} & \textbf{50.13} & \textbf{55.21} & \textbf{75.21} & \textbf{81.42} \\
\midrule
\multicolumn{7}{l}{\textit{End-to-End Systems on WideSearch-zh}} \\
Claude & 0.00 & 0.00 & 20.84 & 28.92 & 43.51 & 52.14 \\
Gemini & 1.50 & 4.00 & 32.32 & 40.52 & 52.92 & 60.44 \\
OpenAI o3 & 3.00 & 5.00 & 27.40 & 38.34 & 46.03 & 56.51 \\
\midrule
\multicolumn{7}{l}{\textit{End-to-End Systems on WideSearch-en}} \\
Claude & 5.00 & 10.00 & 27.39 & 38.07 & 53.29 & 64.81 \\
Gemini & 7.00 & 12.00 & 40.95 & 50.29 & 65.18 & 73.90 \\
OpenAI o3 & 3.00 & 6.00 & 20.42 & 33.72 & 45.02 & 56.47 \\
\midrule
\end{tabular}
}
\end{table*}

%% file: figures/case_1.tex
\begingroup
\centering
\small

    \begin{tcolorbox}[ casebox=yellow!5, title=1. User Request (Task $Q$)]
        Could you provide a detailed list of Michelin three-star restaurants in Paris, France as of December 31, 2024? I especially want to know the name, main cuisine style and exact address of each restaurant. Please organise the results in one Markdown table with the following columns: Restaurant, Main Cuisine Style, Address. Don't ask me any questions, just output the results according to the column without omitting cells arbitrarily. 
        The output format is "markdown \textbackslash n\{data\_content\}\textbackslash n"
    \end{tcolorbox}

    \begin{tcolorbox}[casebox=blue!5, title=2. Host Agent ($\hat{A}$)]
        \textbf{Planning Stage 1 ($q_0$):} Search for Official Michelin three-star restaurants in Paris, France as of December 31, 2024, from reliable sources like Wikipedia.
    \end{tcolorbox}

    \begin{tcolorbox}[casebox=green!5, title=3. Manager Agent ($\tilde{A}$)]
        \textbf{Subtask 1 ($q^0_0$):} Official Michelin Guide list of three Michelin star restaurants in Paris as of 2024 (page or press release listing all Paris 3-star restaurants, include restaurant names and addresses). \\\\
        \textbf{Subtask 2 ($q^1_0$):} Wikipedia page listing Michelin three-starred restaurants in Paris as of 2024 (list and addresses). 
    \end{tcolorbox}

    \begin{tcolorbox}[casebox=red!5, title=4. Worker Agents ($\{\bar{A}\}$)]
        \textbf{Worker 1 Return ($\bar{A}_0^0(q_0^0)$):} Source (official): MICHELIN Guide - Paris restaurants with three MICHELIN stars: L'Ambroisie - [Loc.] [URL]; \'{E}picure (Le Bristol) [Loc.] [URL] [...]. \\\\
        \textbf{Worker 2 Return ($\bar{A}_0^1(q_0^1)$):} Here are the Michelin three‑star restaurants located in Paris listed on Wikipedia (as of the 2024 Michelin Guide):\textbackslash n\textbackslash n \textbackslash n\textbackslash n- Kei (1st arrondissement) — Kei Kobayashi \textbackslash n\textbackslash n- Plénitude — Cheval Blanc Paris […]
    \end{tcolorbox}

    \begin{tcolorbox}[casebox=green!5, title=5. Manager Agent ($\tilde{A}$)]
        \textbf{Manager Return ($\text{Aggr}_{\tilde{A} _{0}}\!\Big(
\{\bar{A}^{k}_{0}(q^{k}_{0})\}_{k=0}^{1}
\Big)$):} Source (official): MICHELIN Guide - Paris restaurants with three MICHELIN stars: L'Ambroisie - [Loc.] [URL]; \'{E}picure (Le Bristol) [Loc.] [URL] [...].
\textbackslash n Source: Wikipedia: Kei (1st arrondissement) — Kei Kobayashi; Plénitude — Cheval Blanc Paris […]
    \end{tcolorbox}

    \begin{tcolorbox}[ casebox=blue!5, title=6. Host Agent ($\hat{A}$)]
        \textbf{Planning Stage 2 ($q_1$):} Search for main cuisine style and exact street address for restaurants: L'Ambroisie Paris, Épicure (Le Bristol Paris), […] as of Dec 31, 2024. Use Michelin Guide or official site.
    \end{tcolorbox}

    \begin{tcolorbox}[ casebox=green!5, title=7. Manager Agent ($\tilde{A}$)]
        \textbf{Subtask 1 ($q^0_1$):} L'Ambroisie Paris — Provide the main cuisine style and exact street address (including postal code) of L'Ambroisie, Paris, as of Dec 31, 2024. Use Michelin Guide or the restaurant's official site. \\\\
        \textbf{Subtasks 2-9 ($q^{1-8}_1$):} [...]\\\\
        \textbf{Subtask 10 ($q^{9}_1$):} Épicure (Le Bristol Paris) — Provide the main cuisine style and exact street address (including postal code) of Épicure, Le Bristol Paris, as of Dec 31, 2024. Use Michelin Guide or official site.
    \end{tcolorbox}
    
    \begin{tcolorbox}[ casebox=red!5, title=8. Worker Agents ($\{\bar{A}\}$)]
        \textbf{Worker 1 Return ($\bar{A}_1^0(q_1^0)$):} Source (official): Cuisine style: Classic French / French haute cuisine (Michelin: \textbackslash\textbackslash"Classic Cuisine\textbackslash\textbackslash")\textbackslash\textbackslash n- Exact address: Place des Vosges, 75004 Paris, France\textbackslash\textbackslash n\textbackslash\textbackslash nSources: Michelin Guide entry for L\'Ambroisie and the restaurant\'s official website (ambroisie-paris.com). \\\\
        \textbf{Workers 2-9 Return ($\bar{A}_1^{1-8}(q_1^{1-8})$):}[...] \\\\
        \textbf{Worker 10 Return ($\bar{A}_1^1(q_1^1)$):} Épicure (Le Bristol Paris)\textbackslash\textbackslash n- Main cuisine style: Modern cuisine (French fine dining)\textbackslash\textbackslash n- Exact street address (as of 31 Dec 2024): Le Bristol, 112 Rue du Faubourg-Saint-Honoré, 75008 Paris, France\textbackslash\textbackslash n\textbackslash\textbackslash nSources: MICHELIN Guide entry for Épicure and Le Bristol Paris official site (Oetker Collection).
    \end{tcolorbox}

    \begin{tcolorbox}[casebox=green!5, title=9. Manager Agent ($\tilde{A}$)]
        \textbf{Manager Return ($\text{Aggr}_{\tilde{A} _{1}}\!\Big(
\{\bar{A}^{k}_{1}(q^{k}_{1})\}_{k=0}^{9}
\Big)$):} L'Ambroisie Paris - Cuisine style: Classic French / French haute cuisine (Michelin: \textbackslash"Classic Cuisine\textbackslash")\textbackslash n- Exact address: Place des Vosges, 75004 Paris, France \textbackslash n \textbackslash\textbackslash nSources: Michelin Guide entry for L\'Ambroisie and the restaurant\'s official website (ambroisie-paris.com). 
\textbackslash n\textbackslash n Épicure (Le Bristol Paris)\textbackslash\textbackslash n- Main cuisine style: Modern cuisine (French fine dining)\textbackslash\textbackslash n- Exact street address (as of 31 Dec 2024): Le Bristol, 112 Rue du Faubourg-Saint-Honoré, 75008 Paris, France\textbackslash\textbackslash n \textbackslash\textbackslash nSources: MICHELIN Guide entry for Épicure and Le Bristol Paris official site (Oetker Collection). [...] 
    \end{tcolorbox}

    \begin{tcolorbox}[casebox=blue!5, title=10. Host Agent ($\hat{A}$)]
\textbf{Host Return ($y^*$):}

\vspace{0.5em}
\small
\setlength{\tabcolsep}{4pt}
\renewcommand{\arraystretch}{1.15}
\begin{tabular}{|p{3.0cm}|p{5.0cm}|p{6.0cm}|}
\hline
\textbf{Restaurant} & \textbf{Main Cuisine Style} & \textbf{Address} \\
\hline
L'Ambroisie &
French haute cuisine (classic) &
9 Place des Vosges, 75004 Paris, France \\
\hline

\'{E}picure (Le Bristol) &
Modern French / French fine dining &
Le Bristol, 112 Rue du Faubourg-Saint-Honor\'{e}, 75008 Paris, France \\
\hline

Arp\`{e}ge &
Vegetable-focused creative French / haute cuisine &
84 Rue de Varenne, 75007 Paris, France \\
\hline

Le Pr\'{e} Catelan &
Contemporary French / creative haute cuisine &
Route de Suresnes, Bois de Boulogne, 75016 Paris, France \\
\hline

Le Cinq (Four Seasons Hotel George~V) &
Modern French / French fine dining &
Four Seasons Hotel George~V, 31 Avenue George~V, 75008 Paris, France \\
\hline

Kei &
Modern creative (French technique with Japanese influence) &
5 Rue Coq-H\'{e}ron, 75001 Paris, France \\
\hline

Pl\'{e}nitude --- Cheval Blanc Paris &
Creative gastronomic French &
8 Quai du Louvre, 75001 Paris, France \\
\hline

Pierre Gagnaire &
Creative French / modern haute cuisine &
6 Rue Balzac, 75008 Paris, France \\
\hline

All\'{e}no Paris au Pavillon Ledoyen &
Creative French / haute cuisine &
8 Avenue Dutuit, 75008 Paris, France \\
\hline

Le Gabriel (La R\'{e}serve Paris) &
Creative contemporary French &
La R\'{e}serve Paris, 42 Avenue Gabriel, 75008 Paris, France \\
\hline
\end{tabular}
\end{tcolorbox}


\captionof{figure}{Execution trace: Michelin three-star restaurant synthesis.}\label{fig:case_1}

\endgroup

%% file: figures/case_2.tex
\begingroup
\centering
\small
\begin{CJK*}{UTF8}{gbsn}

\begin{tcolorbox}[casebox=yellow!5, breakable, title=1. User Query (Task)]
        在此由史实改编的经典作品中，一名演员饰演的人物在历史中为四朝元老，并且以武入仕，那么此经典作品中的女主角的父亲是？ \texttt{[instance\_id: 169]}
    \end{tcolorbox}

    \begin{tcolorbox}[casebox=blue!5, title=2. Host Agent ($\hat{A}$)]
        \textbf{Reasoning:} 任务涉及历史考据与影视剧情细节。需先锁定历史原型与作品（广度搜索），再精准查证人物关系（深度阅读）。 \\
        \textbf{Plan Step 1 (To SearchManager):} 检索“四朝元老、以武入仕”对应的历史名臣及相关经典影视改编。 \\
        \textbf{Plan Step 2 (To BrowserManager):} 针对锁定的作品（如《醉打金枝》），阅读其百科页面，精准提取女主角的生父信息，排除干扰。
    \end{tcolorbox}

    \begin{tcolorbox}[casebox=green!5, title=3. Manager Layer (Search \& Browser Managers)]
        \textbf{[SearchManager] Decomposition:} 
        将 Step 1 拆解为并行查询：(a) “唐朝 四朝元老 武举名将”；(b) “以郭子仪为主角的电视剧”。 \\
        \textbf{[BrowserManager] Decomposition:} 
        基于 Search 结果，构建导航指令：(a) Google 搜索“醉打金枝 维基百科”；(b) 进入条目读取“角色介绍”表格；(c) 验证“升平公主”的父亲字段。
    \end{tcolorbox}

    \begin{tcolorbox}[casebox=red!5, title=4. Worker Layer (Tool Execution)]
        \textbf{[Search Worker] Results:} 
        \begin{itemize}[leftmargin=1.5em, topsep=0pt, itemsep=0pt]
            \item \textit{History:} 锁定\textbf{郭子仪}（历仕玄肃代德四朝，武举异等）。
            \item \textit{Drama:} 锁定作品\textbf{《醉打金枝》}。
        \end{itemize}
        \textbf{[Browser Worker] Results:} 
        \begin{itemize}[leftmargin=1.5em, topsep=0pt, itemsep=0pt]
            \item \textit{Navigation:} 成功加载《醉打金枝》词条页面。
            \item \textit{Extraction:} 读取到女主角“升平公主”是唐代宗的女儿（\textbf{李豫}），而郭子仪是其公公（郭暧之父）。
        \end{itemize}
    \end{tcolorbox}

    \begin{tcolorbox}[casebox=violet!5, title=5. Final Output]
        \textbf{Answer Generation:} \textbf{唐代宗（李豫）}
    \end{tcolorbox}

\end{CJK*}

\par\medskip
\phantomsection
\captionsetup{hypcap=false}
\captionof{figure}{Execution trace: Historical riddle reasoning.}\label{fig:case_2}
\endgroup

%% file: figures/failure_case.tex
\begin{CJK*}{UTF8}{gbsn}
\newpage
\begin{center}
\footnotesize
\end{center}

\begin{figure*}[h]
\begin{tcolorbox}[casebox=yellow!5, title=1. User Query (Task)]
某种遗传性疾病，存在一种变异型，此疾病很多患者伴有其他病变，患者常表现有某种多发于女性的常染色体显性疾病的症状。
该疾病为常染色体疾病，大部分患者死亡年龄较早，请问该疾病名称？
\texttt{[instance\_id: 126]}
\end{tcolorbox}

\begin{tcolorbox}[casebox=blue!5, title=2. Host Agent ($\hat{A}$)]
\textbf{Reasoning:} 题干线索抽象且包含“变异型”“多系统病变”“早亡”等特征，需通过检索定位候选疾病名称并完成消歧。\\
\textbf{Plan (To SearchManager):} 检索并锁定最可能的\emph{标准疾病名称}（单一实体），而非输出鉴别诊断列表。
\end{tcolorbox}

\begin{tcolorbox}[casebox=green!5, title=3. Manager Layer (SearchManager)]
\textbf{Decomposition:}
(a) 将“变异型”作为\emph{已命名亚型}线索检索；(b) 交叉验证“常染色体遗传 + 多系统受累 + 早亡”的候选；
(c) 若不唯一，优先匹配题库/数据集使用的\emph{规范中文病名}。
\end{tcolorbox}

\begin{tcolorbox}[casebox=red!5, title=4. Worker Layer (Observed Evidence)]
\textbf{Observed outcome:} 未能检索到唯一标准命名实体，转而生成“相近候选”列表，并将“变异型”解释为广义表型/基因变异。\\
\textbf{Top candidate selected:} 变异型转甲状腺素蛋白淀粉样变性（ATTRv / TTR 相关家族性淀粉样变性）。
\end{tcolorbox}

\begin{tcolorbox}[casebox=violet!5, title=5. Output vs.\ Gold Answer (Failure)]
\textbf{System predicted:} 变异型转甲状腺素蛋白淀粉样变性（TTR 相关家族性淀粉样变性）\\
\textbf{Gold answer:} 着色性干皮病\\
\textbf{Outcome:} \textbf{Incorrect} (counted as a failure).
\end{tcolorbox}

\captionof{figure}{\textbf{Failure case from BrowseComp-zh.}
\emph{Gold}: \textit{Xeroderma pigmentosum} (着色性干皮病). \emph{Predicted}: variant transthyretin amyloidosis (ATTRv / TTR 相关家族性淀粉样变性).
This failure arises from an \textbf{answer-type mismatch}: the benchmark expects a single canonical disease entity (entity linking), but the system returned a plausible disease class after treating “变异型” as a generic notion of genetic/phenotypic variation rather than a subtype-naming cue, and then selecting the most plausible candidate under uncertainty.}
\label{fig:failure_case_126}

\end{figure*}

\end{CJK*}

%% file: figures/failure.tex
\begin{CJK*}{UTF8}{gbsn}
\newpage
\begin{center}
\footnotesize
\end{center}

\begin{figure*}[h]
\begin{tcolorbox}[casebox=yellow!5, title=1. User Query (Task)]
It is said that "Mom Su" has saved AMD. Please help me sort out the specific processor (CPU) products released by AMD and their detailed information during the decade since Lisa Su became the CEO of AMD to 2023 (include 2023)...
By the way, the core architecture of processor(CPU) should be Zen...
Please output the sorted data in the format of one Markdown table... Don't ask me any questions, just output the results according to the column without omitting cells arbitrarily.
\texttt{[instance\_id: ws\_en\_091]}
\end{tcolorbox}

\begin{tcolorbox}[casebox=blue!5, title=2. Host Agent ($\hat{A}$)]
\textbf{Reasoning:} The user requests a \emph{complete, sorted database} of all AMD Zen CPUs (Desktop, Mobile, Server) from ~2014–2023 with 12 specific technical columns. This requires aggregating hundreds of SKUs.\\
\textbf{Plan (To SearchManager):} Execute parallel searches to gather full SKU lists for Zen, Zen+, Zen 2, Zen 3, and Zen 4 architectures, then compile into a single Markdown table.
\end{tcolorbox}

\begin{tcolorbox}[casebox=green!5, title=3. Manager Layer (SearchManager)]
\textbf{Decomposition:}
(a) Retrieve Desktop/HEDT Zen CPU lists; (b) Retrieve Mobile Zen CPU lists; (c) Retrieve APU lists; (d) Identify official specs for cross-checking.\\
\textbf{Execution Issue:} The subtask for retrieving the APU list resulted in a massive context payload (scraping entire tables for hundreds of models).
\end{tcolorbox}

\begin{tcolorbox}[casebox=red!5, title=4. Worker Layer (Observed Evidence)]
\textbf{Error Encountered:} \texttt{BadRequestError: Input tokens exceed limit of 272000 tokens. Resulted in 299751 tokens.}\\
\textbf{Agent Adaptation:} The agent recognized it could not strictly fulfill the "do not omit cells" requirement due to technical limits. It pivoted to providing a \emph{"representative sample"} table (approx. 18 rows) and a tutorial on how the user could generate the full list themselves using TechPowerUp.
\end{tcolorbox}

\begin{tcolorbox}[casebox=violet!5, title=5. Output vs.\ Required Answer (Failure)]
\textbf{System predicted:} A table with ~18 "representative" rows + a disclaimer: \emph{"For a full exhaustive list... please export from TechPowerUp..."}\\
\textbf{Gold Requirement:} A complete table containing \emph{all} specific processor products released in that timeframe (hundreds of rows).\\
\textbf{Outcome:} \textbf{Incomplete} (Failure due to context constraints).
\end{tcolorbox}

\captionof{figure}{\textbf{Failure case from WideSearch (ws\_en\_091).}
\emph{Requirement}: Comprehensive aggregation of all AMD Zen CPUs (2017--2023). \emph{Predicted}: A sampled subset.
This failure arises from \textbf{Context Length Constraints / Scalability limits}. The agent successfully identified the data sources but failed to process the volume of data (hundreds of SKUs $\times$ 12 columns) within the LLM's context window (hitting ~300k tokens), forcing a fallback to an incomplete answer despite the user's explicit instruction not to omit data.}
\label{fig:failure_case_ws_091}

\end{figure*}

\end{CJK*}

%% file: main.bib
@article{dean2008mapreduce,
  title={MapReduce: simplified data processing on large clusters},
  author={Dean, Jeffrey and Ghemawat, Sanjay},
  journal={Communications of the ACM},
  volume={51},
  number={1},
  pages={107--113},
  year={2008},
  publisher={ACM New York, NY, USA}
}

@article{hu2025owl,
  title={Owl: Optimized workforce learning for general multi-agent assistance in real-world task automation},
  author={Hu, Mengkang and Zhou, Yuhang and Fan, Wendong and Nie, Yuzhou and Xia, Bowei and Sun, Tao and Ye, Ziyu and Jin, Zhaoxuan and Li, Yingru and Chen, Qiguang and others},
  journal={arXiv preprint arXiv:2505.23885},
  year={2025}
}

@inproceedings{
mialon2023gaia,
title={{GAIA}: a benchmark for General {AI} Assistants},
author={Gr{\'e}goire Mialon and Cl{\'e}mentine Fourrier and Thomas Wolf and Yann LeCun and Thomas Scialom},
booktitle={The Twelfth International Conference on Learning Representations},
year={2024},
url={https://openreview.net/forum?id=fibxvahvs3}
}

@article{nakano2021webgpt,
  title={Webgpt: Browser-assisted question-answering with human feedback},
  author={Nakano, Reiichiro and Hilton, Jacob and Balaji, Suchir and Wu, Jeff and Ouyang, Long and Kim, Christina and Hesse, Christopher and Jain, Shantanu and Kosaraju, Vineet and Saunders, William and others},
  journal={arXiv preprint arXiv:2112.09332},
  year={2021}
}

@article{wei2025browsecomp,
  title={Browsecomp: A simple yet challenging benchmark for browsing agents},
  author={Wei, Jason and Sun, Zhiqing and Papay, Spencer and McKinney, Scott and Han, Jeffrey and Fulford, Isa and Chung, Hyung Won and Passos, Alex Tachard and Fedus, William and Glaese, Amelia},
  journal={arXiv preprint arXiv:2504.12516},
  year={2025}
}

@article{wong2025widesearch,
  title={Widesearch: Benchmarking agentic broad info-seeking},
  author={Wong, Ryan and Wang, Jiawei and Zhao, Junjie and Chen, Li and Gao, Yan and Zhang, Long and Zhou, Xuan and Wang, Zuo and Xiang, Kai and Zhang, Ge and others},
  journal={arXiv preprint arXiv:2508.07999},
  year={2025}
}

@inproceedings{wu2024autogen,
  title={Autogen: Enabling next-gen LLM applications via multi-agent conversations},
  author={Wu, Qingyun and Bansal, Gagan and Zhang, Jieyu and Wu, Yiran and Li, Beibin and Zhu, Erkang and Jiang, Li and Zhang, Xiaoyun and Zhang, Shaokun and Liu, Jiale and others},
  booktitle={First Conference on Language Modeling},
  year={2024}
}

@inproceedings{
yao2022react,
title={ReAct: Synergizing Reasoning and Acting in Language Models},
author={Shunyu Yao and Jeffrey Zhao and Dian Yu and Nan Du and Izhak Shafran and Karthik R Narasimhan and Yuan Cao},
booktitle={The Eleventh International Conference on Learning Representations },
year={2023},
url={https://openreview.net/forum?id=WE_vluYUL-X}
}

@article{zhou2025browsecomp,
  title={Browsecomp-ZH: Benchmarking web browsing ability of large language models in chinese},
  author={Zhou, Peilin and Leon, Bruce and Ying, Xiang and Zhang, Can and Shao, Yifan and Ye, Qichen and Chong, Dading and Jin, Zhiling and Xie, Chenxuan and Cao, Meng and others},
  journal={arXiv preprint arXiv:2504.19314},
  year={2025}
}

@article{huang2025deep,
  title={Deep research agents: A systematic examination and roadmap},
  author={Huang, Yuxuan and Chen, Yihang and Zhang, Haozheng and Li, Kang and Zhou, Huichi and Fang, Meng and Yang, Linyi and Li, Xiaoguang and Shang, Lifeng and Xu, Songcen and others},
  journal={arXiv preprint arXiv:2506.18096},
  year={2025}
}

@misc{modelcontextprotocol,
  author       = {Soria Parra, David and Spahr-Summers, Justin},
  title        = {Model Context Protocol},
  year         = {2025},
  howpublished = {\url{https://github.com/modelcontextprotocol/modelcontextprotocol}},
  note         = {GitHub repository}
}

@misc{openai2025deepresearch,
  author       = {{OpenAI}},
  title        = {Introducing Deep Research},
  year         = {2025},
  url          = {https://openai.com/zh-Hans-CN/index/introducing-deep-research/},
  note         = {Accessed: 2026-01-01}
}

@misc{google2025deepresearch,
  author       = {{Google}},
  title        = {Gemini Deep Research},
  year         = {2025},
  url          = {https://gemini.google/overview/deep-research/},
  note         = {Accessed: 2026-01-01}
}

@article{gao2023rag,
  author  = {Yunfan Gao and Yun Xiong and Xinyu Gao and Kangxiang Jia
             and Jinliu Pan and Yuxi Bi and Yixin Dai and Jiawei Sun
             and Haofen Wang},
  title   = {Retrieval-augmented generation for large language models: A survey},
  journal = {arXiv preprint arXiv:2312.10997},
  year    = {2023},
}

@inproceedings{
press2022selfask,
title={Measuring and Narrowing the Compositionality Gap in Language Models},
author={Ofir Press and Muru Zhang and Sewon Min and Ludwig Schmidt and Noah A. Smith and Mike Lewis},
booktitle={The 2023 Conference on Empirical Methods in Natural Language Processing},
year={2023},
url={https://openreview.net/forum?id=feiAVaSXdb}
}

@article{khattab2022dsp,
  author  = {Omar Khattab and Keshav Santhanam and Xiang Lisa Li
             and David Hall and Percy Liang and Christopher Potts
             and Matei Zaharia},
  title   = {Demonstrate-Search-Predict: Composing retrieval and language models for knowledge-intensive NLP},
  journal = {arXiv preprint arXiv:2212.14024},
  year    = {2022},
}

@inproceedings{
chen2024mindsearch,
title={MindSearch: Mimicking Human Minds Elicits Deep {AI} Searcher},
author={Zehui Chen and Kuikun Liu and Qiuchen Wang and Jiangning Liu and Wenwei Zhang and Kai Chen and Feng Zhao},
booktitle={The Thirteenth International Conference on Learning Representations},
year={2025},
url={https://openreview.net/forum?id=xgtXkyqw1f}
}

@inproceedings{zhang2025webpilot,
  title={Webpilot: A versatile and autonomous multi-agent system for web task execution with strategic exploration},
  author={Zhang, Yao and Ma, Zijian and Ma, Yunpu and Han, Zhen and Wu, Yu and Tresp, Volker},
  booktitle={Proceedings of the AAAI Conference on Artificial Intelligence},
  volume={39},
  number={22},
  pages={23378--23386},
  year={2025}
}

@inproceedings{
chan2024rq,
title={{RQ}-{RAG}: Learning to Refine Queries for Retrieval Augmented Generation},
author={Chi-Min Chan and Chunpu Xu and Ruibin Yuan and Hongyin Luo and Wei Xue and Yike Guo and Jie Fu},
booktitle={First Conference on Language Modeling},
year={2024},
url={https://openreview.net/forum?id=tzE7VqsaJ4}
}

@article{madaan2023self,
  title={Self-refine: Iterative refinement with self-feedback},
  author={Madaan, Aman and Tandon, Niket and Gupta, Prakhar and Hallinan, Skyler and Gao, Luyu and Wiegreffe, Sarah and Alon, Uri and Dziri, Nouha and Prabhumoye, Shrimai and Yang, Yiming and others},
  journal={Advances in Neural Information Processing Systems},
  volume={36},
  pages={46534--46594},
  year={2023}
}

@article{ahluwalia2024hybrid,
  title={Hybrid Semantic Search: Unveiling User Intent Beyond Keywords},
  author={Ahluwalia, Aman and Sutradhar, Bishwajit and Ghosh, Karishma and Yadav, Indrapal and Sheetal, Arpan and Patil, Prashant},
  journal={arXiv preprint arXiv:2408.09236},
  year={2024}
}

@misc{gpt_researcher,
  author = {Assaf Elovic},
  title = {GPT Researcher: An autonomous agent designed for comprehensive online research},
  year = {2023},
  publisher = {GitHub},
  journal = {GitHub repository},
  howpublished = {\url{https://github.com/assafelovic/gpt-researcher}},
  commit = {main}
}

@misc{open_deep_research,
  author = {{LangChain AI}},
  title = {Open Deep Research: An open source implementation of deep research agents using LangGraph},
  year = {2025},
  publisher = {GitHub},
  journal = {GitHub repository},
  howpublished = {\url{https://github.com/langchain-ai/open_deep_research}},
  note = {Accessed: 2026-01-02}
}

@article{coelho2025deepresearchgym,
  title={Deepresearchgym: A free, transparent, and reproducible evaluation sandbox for deep research},
  author={Coelho, Jo{\~a}o and Ning, Jingjie and He, Jingyuan and Mao, Kangrui and Paladugu, Abhijay and Setlur, Pranav and Jin, Jiahe and Callan, Jamie and Magalh{\~a}es, Jo{\~a}o and Martins, Bruno and others},
  journal={arXiv preprint arXiv:2505.19253},
  year={2025}
}

@inproceedings{
zhang2024aflow,
title={{AF}low: Automating Agentic Workflow Generation},
author={Jiayi Zhang and Jinyu Xiang and Zhaoyang Yu and Fengwei Teng and Xiong-Hui Chen and Jiaqi Chen and Mingchen Zhuge and Xin Cheng and Sirui Hong and Jinlin Wang and Bingnan Zheng and Bang Liu and Yuyu Luo and Chenglin Wu},
booktitle={The Thirteenth International Conference on Learning Representations},
year={2025},
url={https://openreview.net/forum?id=z5uVAKwmjf}
}

@inproceedings{
niu2025flow,
title={Flow: Modularized Agentic Workflow Automation},
author={Boye Niu and Yiliao Song and Kai Lian and Yifan Shen and Yu Yao and Kun Zhang and Tongliang Liu},
booktitle={The Thirteenth International Conference on Learning Representations},
year={2025},
url={https://openreview.net/forum?id=sLKDbuyq99}
}

@misc{langgraph,
  author = {{LangChain AI}},
  title = {LangGraph: Building stateful, multi-actor applications with LLMs},
  year = {2024},
  publisher = {GitHub},
  journal = {GitHub repository},
  howpublished = {\url{https://github.com/langchain-ai/langgraph}},
  note = {Version: latest}
}

@inproceedings{leviathan2023fast,
  title={Fast inference from transformers via speculative decoding},
  author={Leviathan, Yaniv and Kalman, Matan and Matias, Yossi},
  booktitle={International Conference on Machine Learning},
  pages={19274--19286},
  year={2023},
  organization={PMLR}
}

@article{miao2023specinfer,
  title={Specinfer: Accelerating generative large language model serving with tree-based speculative inference and verification},
  author={Miao, Xupeng and Oliaro, Gabriele and Zhang, Zhihao and Cheng, Xinhao and Wang, Zeyu and Zhang, Zhengxin and Wong, Rae Ying Yee and Zhu, Alan and Yang, Lijie and Shi, Xiaoxiang and others},
  journal={arXiv preprint arXiv:2305.09781},
  year={2023}
}

@inproceedings{
cai2024medusa,
title={Medusa: Simple {LLM} Inference Acceleration Framework with Multiple Decoding Heads},
author={Tianle Cai and Yuhong Li and Zhengyang Geng and Hongwu Peng and Jason D. Lee and Deming Chen and Tri Dao},
booktitle={Forty-first International Conference on Machine Learning},
year={2024},
url={https://openreview.net/forum?id=PEpbUobfJv}
}

@inproceedings{
pan2025specreason,
title={SpecReason: Fast and Accurate Inference-Time Compute via Speculative Reasoning},
author={Rui Pan and Yinwei Dai and Zhihao Zhang and Gabriele Oliaro and Zhihao Jia and Ravi Netravali},
booktitle={NeurIPS 2025 Workshop on Efficient Reasoning},
year={2025},
url={https://openreview.net/forum?id=UrZgP5DD70}
}

@inproceedings{
yang2025speculative,
title={Speculative Thinking: Enhancing Small-Model Reasoning with Large Model Guidance at Inference Time},
author={Van Yang and Xiang Yue and Vipin Chaudhary and Xiaotian Han},
booktitle={Second Conference on Language Modeling},
year={2025},
url={https://openreview.net/forum?id=4Ns18bSoHo}
}

@article{ding2025dynamic,
  title={Dynamic parallel tree search for efficient llm reasoning},
  author={Ding, Yifu and Jiang, Wentao and Liu, Shunyu and Jing, Yongcheng and Guo, Jinyang and Wang, Yingjie and Zhang, Jing and Wang, Zengmao and Liu, Ziwei and Du, Bo and others},
  journal={arXiv preprint arXiv:2502.16235},
  year={2025}
}

@article{wen2025parathinker,
  title={Parathinker: Native parallel thinking as a new paradigm to scale llm test-time compute},
  author={Wen, Hao and Su, Yifan and Zhang, Feifei and Liu, Yunxin and Liu, Yunhao and Zhang, Ya-Qin and Li, Yuanchun},
  journal={arXiv preprint arXiv:2509.04475},
  year={2025}
}

@article{zheng2025parallel,
  title={Parallel-r1: Towards parallel thinking via reinforcement learning},
  author={Zheng, Tong and Zhang, Hongming and Yu, Wenhao and Wang, Xiaoyang and Dai, Runpeng and Liu, Rui and Bao, Huiwen and Huang, Chengsong and Huang, Heng and Yu, Dong},
  journal={arXiv preprint arXiv:2509.07980},
  year={2025}
}

@article{qin2025flash,
  title={Flash-searcher: Fast and effective web agents via dag-based parallel execution},
  author={Qin, Tianrui and Chen, Qianben and Wang, Sinuo and Xing, He and Zhu, King and Zhu, He and Shi, Dingfeng and Liu, Xinxin and Zhang, Ge and Liu, Jiaheng and others},
  journal={arXiv preprint arXiv:2509.25301},
  year={2025}
}

@article{nie2025flashresearch,
  title={FlashResearch: Real-time Agent Orchestration for Efficient Deep Research},
  author={Nie, Lunyiu and Lipka, Nedim and Rossi, Ryan A and Chaudhuri, Swarat},
  journal={arXiv preprint arXiv:2510.05145},
  year={2025}
}

@article{zhao2023survey,
  title={A survey of large language models},
  author={Zhao, Wayne Xin and Zhou, Kun and Li, Junyi and Tang, Tianyi and Wang, Xiaolei and Hou, Yupeng and Min, Yingqian and Zhang, Beichen and Zhang, Junjie and Dong, Zican and others},
  journal={arXiv preprint arXiv:2303.18223},
  volume={1},
  number={2},
  year={2023}
}

@article{yang2025agentic,
  title={Agentic web: Weaving the next web with ai agents},
  author={Yang, Yingxuan and Ma, Mulei and Huang, Yuxuan and Chai, Huacan and Gong, Chenyu and Geng, Haoran and Zhou, Yuanjian and Wen, Ying and Fang, Meng and Chen, Muhao and others},
  journal={arXiv preprint arXiv:2507.21206},
  year={2025}
}

@article{team2025mirothinker,
  title={MiroThinker: Pushing the Performance Boundaries of Open-Source Research Agents via Model, Context, and Interactive Scaling},
  author={Team, MiroMind and Bai, Song and Bing, Lidong and Chen, Carson and Chen, Guanzheng and Chen, Yuntao and Chen, Zhe and Chen, Ziyi and Dai, Jifeng and Dong, Xuan and others},
  journal={arXiv preprint arXiv:2511.11793},
  year={2025}
}

@article{li2025websailor,
  title={WebSailor: Navigating Super-human Reasoning for Web Agent},
  author={Li, Kuan and Zhang, Zhongwang and Yin, Huifeng and Zhang, Liwen and Ou, Litu and Wu, Jialong and Yin, Wenbiao and Li, Baixuan and Tao, Zhengwei and Wang, Xinyu and others},
  journal={arXiv preprint arXiv:2507.02592},
  year={2025}
}

@Inbook{Simon1991,
author="Simon, Herbert A.",
title="The Architecture of Complexity",
bookTitle="Facets of Systems Science",
year="1991",
publisher="Springer US",
address="Boston, MA",
pages="457--476",
isbn="978-1-4899-0718-9",
doi="10.1007/978-1-4899-0718-9_31",
url="https://doi.org/10.1007/978-1-4899-0718-9_31"
}
